\documentclass[10pt,twocolumn,letterpaper]{article}

\usepackage{wacv}              

\usepackage{graphicx} 
\usepackage{tabularx} 
\usepackage{wrapfig}
\usepackage{url}            
\usepackage{booktabs}       
\usepackage{amsfonts}       
\usepackage{nicefrac}       
\usepackage{microtype}      
\usepackage[dvipsnames,table,xcdraw]{xcolor}        
\usepackage{adjustbox}
\usepackage{mathtools}
\usepackage{stfloats} 
\usepackage{tabularx}

\usepackage{array}
\newcolumntype{H}{>{\setbox0=\hbox\bgroup}c<{\egroup}@{}}
\usepackage{float}
\usepackage[skip=5pt, belowskip=-10pt]{caption}
\usepackage[accsupp]{axessibility}

\usepackage[pagebackref,breaklinks,colorlinks]{hyperref}
\usepackage{orcidlink}

\usepackage[capitalize]{cleveref}
\crefname{section}{Sec.}{Secs.}
\Crefname{section}{Section}{Sections}
\Crefname{table}{Table}{Tables}
\crefname{table}{Tab.}{Tabs.}

\newcommand{\sbs}[1]{\leavevmode#1}
\newcommand{\superaffil}[2]{\textsuperscript{#1}\,#2}
\newcommand{\superaffi}[2]{\textsuperscript{#1}\,\\#2}

\def\wacvPaperID{1579} 
\def\confName{WACV}
\def\confYear{2025}

\begin{document}

\title{\texttt{FASTER}: A Font-Agnostic Scene Text Editing and Rendering framework}

\author{
\small Alloy Das \superaffil{1}\thanks{These authors have contributed equally to the work.}\and
\small Sanket Biswas\superaffil{2}\footnotemark[1]\and
\small Prasun Roy\superaffi{3}\and
\small Subhankar Ghosh\superaffi{3}\and
\small Umapada Pal\superaffil{1}\and 
\small Michael Blumenstein\superaffi{3}\and
\small Josep Llad\'{o}s\superaffil{2}\and
\small Saumik Bhattacharya\superaffi{4}\and
\and
\footnotesize
  \small \textsuperscript{1}CVPRU, Indian Statistical Institute, Kolkata
  \quad
  \textsuperscript{2}CVC, Universitat Autònoma de Barcelona
  \\ 
  \quad \small \textsuperscript{3}FEIT, University of Technology Sydney, Australia
  \quad
  \textsuperscript{4}ECE, Indian Institute of Technology, Kharagpur
}
\maketitle

\begin{figure*}[b] 
\centering
\includegraphics[width=\textwidth]{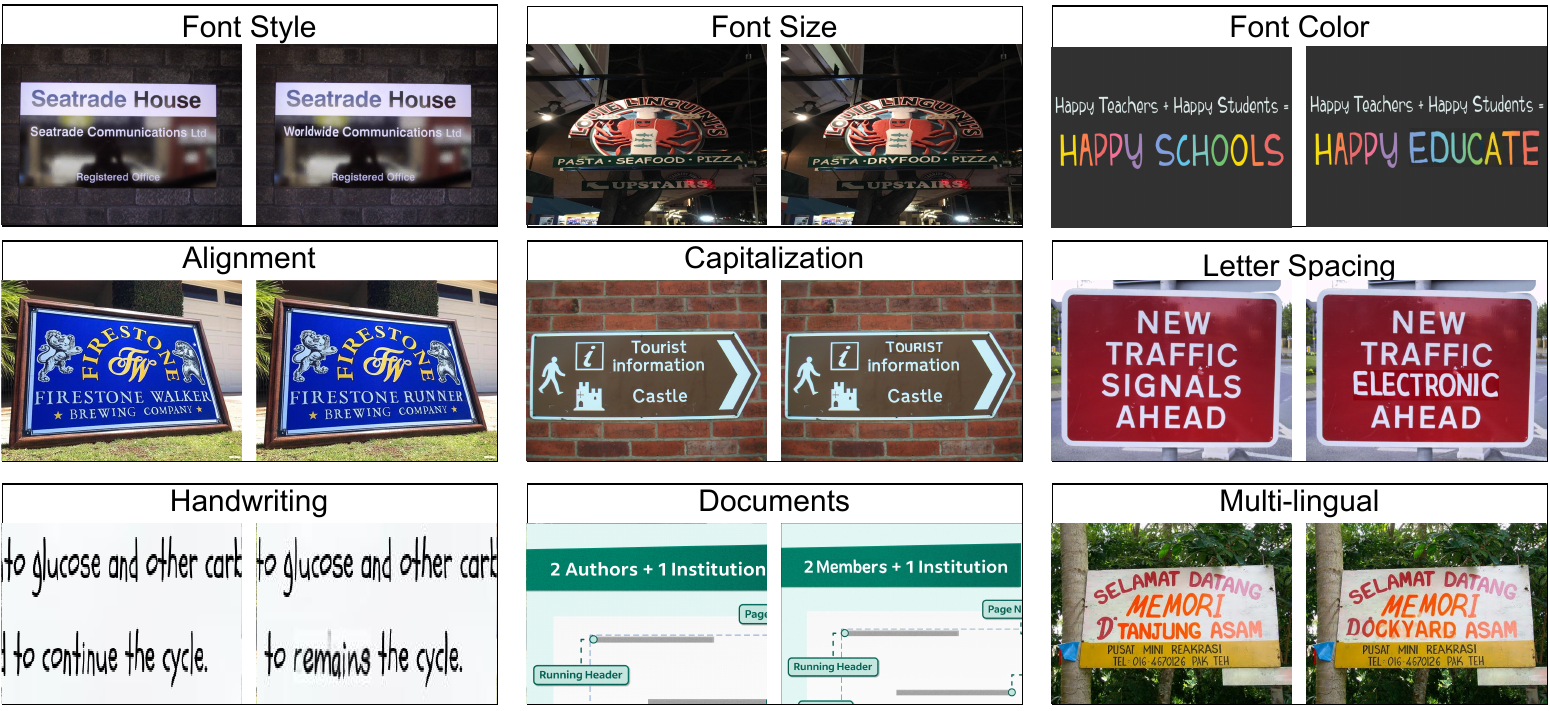}
\caption{Given an \textbf{image} and the \textbf{desired text} to render, FASTER performs appropriate edits on the target text regions in complex real scenes with high consistency and realism on a wide range of typefaces and multiple font attributes with varying word lengths.} \vspace{2em} \label{fig: teaser}
\end{figure*}

\begin{abstract}
Scene Text Editing (STE) is a challenging research problem, that primarily aims towards modifying existing texts in an image while preserving the background and the font style of the original text. Despite its utility in numerous real-world applications, existing style-transfer-based approaches have shown sub-par editing performance due to (1) complex image backgrounds, (2) diverse font attributes, and (3) varying word lengths within the text. To address such limitations, in this paper, we propose a novel \textbf{font-agnostic scene text editing and rendering} framework, named \textbf{FASTER}, for simultaneously generating text in arbitrary styles and locations while preserving a natural and realistic appearance and structure. A combined fusion of target mask generation and style transfer units, with a cascaded self-attention mechanism has been proposed to focus on multi-level text region edits to handle varying word lengths. Extensive evaluation on a real-world database with further subjective human evaluation study indicates the superiority of FASTER in both scene text editing and rendering tasks, in terms of model performance and efficiency. Our code will be released upon acceptance.  
\end{abstract}


\section{Introduction}

In user interfaces, advertisements, graphic design, and augmented reality applications, ensuring a pleasing and immersive user experience is paramount. Text is a fundamental component within visual compositions, anchoring meaning, guiding attention and shaping a document image's overall \textit{visual coherence} by adhering to attributes such as font style, size, color, and alignment~\cite{campbell2013coherence}. In recent years, there has been a surging interest in the field of Scene Text Editing (STE) due to its practical applications across multiple tasks which include text image synthesis~\cite{karacan2016learning, qin2021mask, zhan2018verisimilar}, text style transfer~\cite{zhang2018separating, atarsaikhan2017neural, azadi2018multi}, and augmented reality translation~\cite{cao2023mobile, fragoso2011translatar, du2011snap}. Prior approaches~\cite{lee2021rewritenet, park2021multiple, qu2022exploring, roy2020stefann, wu2019editing, yang2020swaptext} have essentially formulated it as a style transfer problem using generative adversarial networks (GANs)~\cite{goodfellow2014generative}. Yet, the pivotal questions that remain unanswered are: 1) How well can STE models execute seamless text editing while retaining various text attributes, such as font style, size, color, alignment, case distinction, and letter spacing, in complex real-world scenes? 2) How does the inference time of these models impact their practical usefulness and efficacy in real-world applications, without compromising the text editing and rendering quality?


Alternative approaches in STE~\cite{huang2022gentext, shi2022aprnet, wu2019editing} often use a reference image to guide text attribute rendering. However, these methods struggle with diverse font attributes and arbitrary geometric transformations due to content and style entanglement. Diffusion-based approaches~\cite{ji2023improving, chen2024diffute} generate high-quality text images but lack precise control over font style, placement, and complex backgrounds. This is due to (1) \textbf{Limited Adaptability and Control}: Focusing on high-level text descriptions~\cite{chen2023textdiffuser} without capturing contextual backgrounds limits adaptability. (2) \textbf{Computational Complexity}: Diffusion methods are computationally intensive, hindering real-time editing. (3) \textbf{Lack of Positional Guidance}: Text rendering requires alignment, spacing, and formatting considerations not handled by diffusion models.

In this work, we propose FASTER, a novel two-stage GAN-based text style transfer architecture. In Stage-I, a style translation block (STB) generates a target style mask. In Stage-II, a content translation block (CTB) conditions text image attributes to generate content over the Stage-I output. This top-down approach improves upon existing designs~\cite{wu2019editing, yang2020swaptext} that often suffer from global information loss and degraded output quality. Unlike TextStyleBrush~\cite{krishnan2023textstylebrush}, which relies heavily on a pre-trained font classifier and text recognizer and struggles with background attributes in complex images, FASTER incorporates a simple and efficient U-Net module~\cite{ronneberger2015u} to preserve the original text appearance by affecting only non-text regions. Additionally, we introduce a cascaded self-attention module in the Stage-I decoder, inspired by prior works~\cite{zhang2019self, Wan_2021_CVPR, pan2022integration}, to model hierarchical, multi-level dependencies across text regions. This decomposition into multiple cascaded layers with sigmoid attention units allows FASTER to dynamically prioritize relevant text parts, enabling faster and more accurate text editing and rendering. Figure~\ref{fig: teaser} illustrates some qualitative results on some complex real-world examples that demonstrate how the precise edits on the target text preserve font (or typeface) characteristics of the text and its layout composition relative to the foreground style. To further support with quantitative evidence, our proposed FASTER design shows a substantial \textit{improvement of \textbf{-23.83} in perceptual FID} over second-best SwapText~\cite{yang2020swaptext} and a notable enhancement in \textit{inference time} (measured in milliseconds (ms) per image) compared to the runner-up SRNet~\cite{wu2019editing}, achieving a \textit{cost reduction of \textbf{15.75} ms}. The major contributions and novelties of this work can be divided into four folds: (1) FASTER is a novel STE framework using a \textit{two-staged style-transfer methodology with positional guidance} for robust text attribute conditioning. (2) The proposed framework integrates a \textit{cascaded self-attention module} in the decoder for efficient scaling and pre-training, enhancing \textit{real-world text-editing capabilities} (e.g., handwritten fonts, documents, multi-lingual scripts) and achieving the fastest inference speed. (3) We also adapt a \textit{novel combination of learning objectives} for the GAN generator which allows high-quality text image rendering without sacrificing model efficiency. (4) A comprehensive evaluation toolkit based on~\cite{qu2022exploring} and human evaluation study has been provided to ensure thorough quantitative and subjective assessments of text editing and rendering quality.

\section{Related Work}\label{sec:relatedwork}

\noindent
\textbf{Text Image Synthesis:}
Image synthesis and rendering have been extensively studied in computer graphics~\cite{debevec2008rendering}. Text image synthesis serves as a data augmentation technique for text identification and detection. Gupta et al.\cite{gupta2016synthetic} developed an engine to produce synthetic text images, while Jaderberg et al.\cite{jaderberg2014synthetic} created a word generator for synthetic word images which had immense influence in domain-generalized scene text spotting tasks~\cite{das2024diving, das2024harnessing, das2024fasttextspotter}. The aim of text image synthesis is to insert text into semantically significant areas of a background image. However, factors such as font size, perspective, and illumination affect the realism of synthesized text images. To address this, Zhan et al.~\cite{zhan2018verisimilar} integrated semantic coherence, visual attention, and adaptive text presentation, achieving visually accurate but still distinguishable synthetic images with limited font options. Recently, Biswas et al.~\cite{Biswas_2024_CVPR} proposed an approach to generate synthetic documents with relevant text content with variable fonts.

To address these restrictions, recent studies have explored GAN-based picture synthesis algorithms. Zhan et al.\cite{zhan2019spatial} introduced a spatial fusion GAN that combines a geometry synthesizer with an appearance synthesizer for realistic synthesis in both geometry and appearance spaces. Yang et al.'s\cite{yang2019controllable} framework allows control over glyph styles using an adjustable parameter, while GA-DAN~\cite{zhan2019ga} models cross-domain shifts in geometry and appearance. MC-GAN~\cite{azadi2018multi} facilitates font style transfer for letters A to Z, and Wu et al.\cite{wu2019editing} developed an end-to-end trainable style retention network for editing text in natural photos. Although diffusion models\cite{dhariwal2021diffusion, rombach2022high, zhang2023adding} have shown impressive generation capabilities, they struggle with precise text rendering. TextDiffuser~\cite{chen2023textdiffuser} addresses this by using a customized text dataset with OCR annotations, creating visually appealing text coherent with backgrounds. However, this approach relies on expensive OCR tools and paired image-text data, making it less ideal for real-world text editing.

\noindent 
\textbf{Text Style Transfer:} The task of transferring an image's visual style from a reference image to a target image is known as image style transfer. Many existing techniques use an encoder-decoder architecture to embed and then decode the input into a desired output. Isola et al.\cite{isola2017image} developed a learnable mapping using aligned image pairings, while Zhu et al.\cite{zhu2017unpaired} introduced cycle consistency loss to generalize mappings for unpaired data. Challenges like creating images from sketches and synthesizing faces have been approached similarly. An algorithm proposed by Yang et al.\cite{yang2017awesome} uses statistical measurements to transfer text effects by modeling distance-based properties. Additionally, Yang et al.\cite{yang2019tet} utilize stylization and de-stylization sub-networks for style transfer and removal. Other works integrate background and text style information for artifact-free text editing~\cite{shimoda2021rendering}, focus on text image generation with typographic attributes~\cite{inoue2023towards, shimoda2023towards}, and explore style and content disentanglement for OCR-based recognition~\cite{kang2021content}, along with attribute-conditioned text-style transfer~\cite{gomez2019selective, krishnan2023textstylebrush,yamaguchi2021canvasvae, Pippi_2023_CVPR}. This work aims to enhance model efficiency in style transfer.

\noindent
\textbf{Scene Text Editing:}
The variety of uses for GAN-based scene text editing has piqued researchers' growing curiosity. To alter a single character, for instance, Roy et al.~\cite{roy2020stefann} created the Font Adaptive Neural Network (STEFANN). This character-level alteration, however, falls short of replacing words with length alterations, which restricts practical uses. In~\cite{wu2019editing}, the authors developed the word-level editing approach employing three sub-networks: background inpainting, text conversion, and fusion to overcome this constraint. Each module can manage a reasonably straightforward task thanks to the divide-and-conquer method.
To make the text conversion module easier to learn,~\cite{yang2020swaptext} enhanced SRNet and added the TPS module, which separates the spatial transformation from text styles. Furthermore,~\cite{zhao2021deep}  suggested the forge-and-recapture procedure to reduce visual artifacts and applied scene text editing to document forging. To keep the consistency of all other edited frames,~\cite{subramanian2021strive}  used SRNet for video text editing, altering only the chosen frame as a reference and applying certain photometric modifications. These methods, which are essentially extensions of SRNet, can only be trained on artificial datasets and may not be able to replace text with complicated styles. A stroke-level alteration technique that creates more readable text graphics are suggested by~\cite{qu2022exploring}. Their technique can be trained on both labeled synthetic datasets and unpaired scene text images, and it supports the semi-supervised training scheme. \sbs{Recently, stable diffusion-based approaches~\cite{chen2024diffute, ji2023improving, nikolaidou2024diffusionpen} have been investigated for text editing in natural scenes and handwriting domain. This work conducts a rigorous comparison with ~\cite{ji2023improving} on the STE task.}

\section{Proposed Methodology}\label{sec:methodology}
\begin{figure*}[t]
  \centering
  \includegraphics[width=\textwidth, height = 11cm]{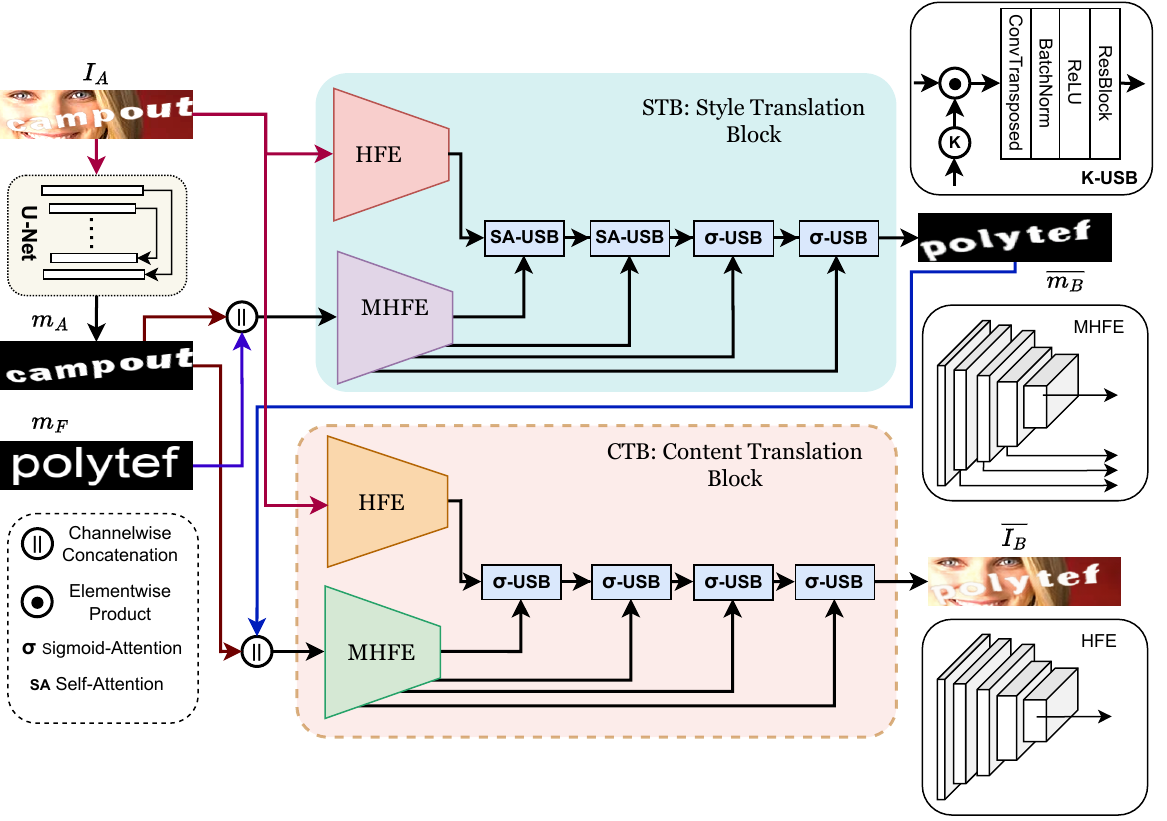}
  \caption{\textbf{Overall Architecture of \texttt{FASTER}}. In \textbf{Stage-I}, an approximate target style mask $\overline{m_B}$ is estimated from the source image $I_A$, source style mask $m_A$, and a fixed style mask $m_F$ of the target text. In \textbf{Stage-II}, the target image $\overline{I_B}$ is generated by transferring image attributes from $I_A$ and conditioning the image translation on the structural guidance $(m_A, \overline{m_B})$.}
  \label{fig:architecture}
\end{figure*}

In this section, we will examine the proposed methodology of FASTER, including the problem formulation, the two stages (task modules) used in the framework, and the optimal learning objectives utilized in the architecture.
\sbs{\subsection{Problem Formulation}}

\noindent
Given a scene text image $I_A$, the objective of the proposed STE is to generate an image $I_B$ with a modified text. To enforce a classifier-free image translation, we aim to condition the generative process on the structural guidance $(m_A, m_B)$, where $m_A$ and $m_B$ correspond to the binary masks of text content in $I_A$ and $I_B$, respectively. However, obtaining $m_B$ before generating $I_B$ is unrealistic, making the end-to-end text style transfer difficult. We address this issue by splitting the generative process into two independent stages containing the \textit{Style Translation Block (STB)} and \textit{Content Translation Block (CTB)}, respectively. \\
\noindent
- In \textit{Stage-I}, we replace the initially unknown mask $m_B$ with another mask $m_F$ having the same textual content but in a fixed font of known style. In this stage, the generator $G_m$ produces an approximation of the target style mask $\overline{m_B}$. \\
\noindent
- In \textit{Stage-II}, an identical generator $G_i$ synthesizes the approximate target image $\overline{I_B}$ by transferring attributes from $I_A$ and using $(m_A, \overline{m_B})$ as structural guidance. We use synthetically generated $(I_A, m_A)$ pairs to train both $G_m$ and $G_i$.\\ 
\noindent
- Additionally, a U-Net~\cite{ronneberger2015u} \textit{feature extraction backbone} inspired from~\cite{roy2023multi} has been separately trained for estimating the mask $\overline{m_A}$ from $I_A$ during inference on real scene text samples. The entire overview of our architecture is shown in Figure~\ref{fig:architecture}. 

\subsection{Stage -- I: Target Style Mask Estimation}
\noindent
The Stage-I architecture includes a functional STB module which is a GAN network containing a target mask generator $G_m$ and a PatchGAN~\cite{isola2017image} discriminator $D_m$. $G_m$ takes the source image $I_A$ and the channel-wise concatenated masks $m_\theta = (m_A, m_F)$ as inputs and produces an approximate target style mask $\overline{m_B}$ as the output. The discriminator $D_m$ discriminates between real and fake transformations by processing channel-wise concatenated masks $(m_A, m_B)$ or $(m_A, \overline{m_B})$, predicting a binary class probability map of ones (real) or zeros (fake).
\\
\noindent
\textbf{Encoder:} $G_m$ comprises two encoding branches for $I_A$ and $m_\theta$ to be referred as \textit{Hierarchical Feature Extractor (HFE)} and \textit{Multi-scale Hierarchical Feature Extractor (MHFE)}. The condition image and the guidance masks are resized to a dimension of $64 \times 256$. At every branch (HFE and MHFE), the encoder first projects the input into a 64-channel feature space using $3 \times 3$ convolution kernels with stride $1$, padding $1$, and without adding any bias. The final block output is obtained after batch normalization (BN) and ReLU units. The projected feature space is then downscaled four times. At every downsampling block, the spatial feature dimension is downscaled to half while doubling the number of channels. Each downsampling block uses $4 \times 4$ convolution kernels, stride $1$, padding $1$, and zero bias. Each downscaling convolution is immediately followed by BN, ReLU and a basic residual block \cite{he2016deep} to generate the final block output.\\
\noindent
\textbf{Decoder:} During decoding, the outputs of both encoding branches HFE and MHFE are channel-wise concatenated and passed through the decoder. The decoder consists of four \textit{UpScaling Blocks (USB)}, each doubling the spatial feature dimension while decreasing the number of channels to half. The upscaling is performed with $4 \times 4$ transposed convolution kernels, stride $1$, padding $1$, and zero bias. Like the encoder, the final block output is generated following BN, ReLU activation, and a basic residual block. In $G_m$, we apply two different attention mechanisms at matching feature resolutions of the encoder and decoder to attend to both coarse and fine image attributes during structural transformation. At the two lowest resolutions, we use \emph{cascaded self-attention} similar to \cite{zhang2019self}, and at the other two higher resolutions, we use a \emph{sigmoid attention}. The self-attention units at the two lowest resolutions $k = \{1, 2\}$ can  represented mathematically as:
\[ m^{\phi_m}_1 = {\phi_m}_1(I^{\varphi_m}_4 \; \odot \; SA(m^{\varphi_m}_4)) \]
\[ m^{\phi_m}_2 = {\phi_m}_2(m^{\phi_m}_1 \; \odot \; SA(m^{\varphi_m}_3)) \]

and for the following decoder blocks at higher resolutions $k = \{3, 4\}$ that use sigmoid attention can be represented as:

\[ m^{\phi_m}_k = {\phi_m}_k(m^{\phi_m}_{k-1} \; \odot \; \sigma(m^{\varphi_m}_{5-k})) \]

where, $I^{\varphi_m}_k$ is the output of $k$-th image encoder block, $m^{\varphi_m}_k$ is the output of $k$-th mask encoder block, $m^{\phi_m}_k$ is the output of the $k$-th decoder block, $SA$ and $\sigma$ denote \emph{self-attention} and \emph{sigmoid attention}, respectively, and $\odot$ denotes element-wise product. The output feature maps from the decoder are post-processed through four consecutive basic residual blocks. The resulting feature space is projected to a 3-channel image space of spatial resolution of $64 \times 256$ by a point convolution with $1 \times 1$ kernel, unit stride, zero padding, and without bias. The final normalized output of $G_m$ is obtained following a hyperbolic tangent $(\mathbf{tanh})$ activation function.

\subsection{Stage -- II: Text Style Transfer with Structural Guidance}
\noindent
The Stage II architecture is identical to Stage-I, with slightly different input specifications and attention mechanisms. In this case, the generator $G_i$ takes the source image $I_A$ and the channel-wise concatenated masks $(m_A, \overline{m_B})$ as inputs and produces an approximate target image $\overline{I_B}$ as the output. The PatchGAN discriminator $D_i$ discerns between a real and a fake transformation by taking channel-wise concatenated images $(I_A, I_B)$ or $(I_A, \overline{I_B})$ and predicting a binary class probability map of ones (real) or zeroes (fake).

In $G_i$, we apply only \emph{sigmoid attention} at every matching feature resolution of the encoder and decoder. Mathematically, at the lowest resolution $k = 1$,

\[ I^{\phi_i}_1 = {\phi_i}_1(I^{\varphi_i}_4 \; \odot \; \sigma(m^{\varphi_i}_4)) \]
\noindent
and for the following decoder blocks at higher resolutions $k = \{2, 3, 4\}$,

\[ I^{\phi_i}_k = {\phi_i}_k(I^{\phi_i}_{k-1} \; \odot \; \sigma(m^{\varphi_i}_{5-k})) \]

where, $I^{\varphi_i}_k$ is the output of $k$-th image encoder block, $m^{\varphi_i}_k$ is the output of $k$-th mask encoder block, $I^{\phi_i}_k$ is the output of the $k$-th decoder block, $\sigma$ denotes \emph{sigmoid attention}, and $\odot$ denotes element-wise product.

\subsection{Learning Objectives}

\noindent
\textbf{Stage -- I Objectives:}
The optimization objective of generator $G_m$ consists of four different loss components -- (a) pixel-wise $L_2$ loss $\mathcal{L}_{L_2}^{G_m}$, (b) discriminator loss $\mathcal{L}_{GAN}^{G_m}$ estimated by the discriminator $D_m$, (c) perceptual loss $\mathcal{L}_{P_{\rho}}^{G_m}$ computed using a pre-trained VGG-19 network \cite{simonyan2014very},  (d) multi-scale structural similarity \cite{wang2004image} loss $\mathcal{L}_{SSIM}^{G_m}$, and (e) OCR perceptual loss~\cite{rodriguez2023ocr} $\mathcal{L}_{{OCR}_{per}}^{G_m}$ computed using pretrained CRAFT~\cite{baek2019character} model.
The $L_2$ reconstruction loss is estimated as the mean squared error (MSE) between the generated style mask $\overline{m_B}$ and target style mask $m_B$.
The GAN discriminator objective is defined as the binary cross-entropy (BCE) estimated by $D_m$ computed between predicted mask $\overline{m_B}$ and the input style mask $m_A$.
The image perceptual loss $\mathcal{L}_{P_{\rho}}^{G_m}$ computed between target style mask $m_B$ and generated style mask  $\overline{m_B}$ follows a similar pattern as ~\cite{johnson2016perceptual}. We include two perceptual loss terms for $\rho = 4$ and $\rho = 9$ in the final objective function. 
\noindent
- The multi-scale structural similarity loss  $\mathcal{L}_{SSIM}^{G_m}$ is estimated similarly as ~\cite{wang2004image} between target style mask $m_B$ and generated style mask  $\overline{m_B}$. \\
\noindent
- The OCR perceptual loss follows a similar pattern as in ~\cite{rodriguez2023ocr} between target style mask $m_B$ and generated style mask  $\overline{m_B}$ is defined as: 
\begin{equation}
  \mathcal{L}_{{OCR}_{per}}^{G_m} =  \sum_{a=1}^{\rho} \frac{1}{h_\rho w_\rho} \sum_{h, w} \left\| (\overline{m_B}_{h,w}^{\rho}) - {m_B}_{h,w}^{\rho} \right\|_2^2
\end{equation}
where $\overline{m_B}_{h,w}^{\rho}$ and ${m_B}_{h,w}^{\rho}$ are the activation map of layer $\rho$ has been taken from the pretrained CRAFT~\cite{baek2019character} model, and $\|\cdot\|_2^2$ denotes the $L_2$ norm (mean squared error).\\
\noindent
- The overall objective function $G_m$ of the STB functional module can be given by:
\begin{equation}
\begin{multlined}
  \mathcal{L}_{G_m} = \lambda_1 \cdot \mathcal{L}_{L_2}^{G_m} + \lambda_2 \cdot \mathcal{L}_{GAN}^{G_m} +  \\ 
  \lambda_3 \cdot (\mathcal{L}_{P_4}^{G_m} + \mathcal{L}_{P_9}^{G_m}) + 
  \lambda_4 \cdot \mathcal{L}_{SSIM}^{G_m} + \lambda_5 \cdot \mathcal{L}_{{OCR}_{per}}^{G_m}
\end{multlined}
\end{equation}
where $\lambda_1$, $\lambda_2$, $\lambda_3$, $\lambda_4$, and $\lambda_5$ are the weighing parameters for respective loss components. \\
\noindent
- The optimization objective of the STB discriminator $D_m$ is given by:
\begin{equation}
\begin{multlined}
  \mathcal{L}_{D_m} = \frac{1}{2} [\mathcal{L}_{BCE} (D_m (m_A, m_B), 1) + \\
  \mathcal{L}_{BCE} (D_m (m_A, \overline{m_B}), 0)] 
\end{multlined}
\end{equation}

\noindent
\textbf{Stage -- II Objectives:} Similarly, the final optimization objective of generator $G_i$ for the CTB functional module consists of four loss components -- (a) pixel-wise $L_1$ loss $\mathcal{L}_{L_1}^{G_i}$ between target image $I_B$ and generated image  $\overline{I_B}$, (b) discriminator loss $\mathcal{L}_{GAN}^{G_i}$ estimated by the discriminator $D_i$ between target image $I_B$ and generated image  $\overline{I_B}$, (c) perceptual loss $\mathcal{L}_{P_{\rho}}^{G_i}$ computed using a pre-trained VGG-19 network between target image $I_B$ and generated image  $\overline{I_B}$, and (d) OCR perceptual loss $\mathcal{L}_{{OCR}_{per}}^{G_i}$ computed using the pretrained CRAFT model between target image $I_B$ and generated image  $\overline{I_B}$.

\noindent
- The overall objective function of the CTB GAN generator $G_i$ is given by:
\begin{equation}
\begin{multlined}
  \mathcal{L}_{G_i} = \beta_1 \cdot \mathcal{L}_{L_2}^{G_i} + \beta_2 \cdot \mathcal{L}_{GAN}^{G_i} + \\
  \beta_3 \cdot (\mathcal{L}_{P_4}^{G_i} + \mathcal{L}_{P_9}^{G_i}) + 
  \beta_4 \cdot (\mathcal{L}_{{OCR}_{per}}^{G_i})
\end{multlined}
\end{equation}
where $\beta_1$, $\beta_2$, $\beta_3$, and $\beta_4$ are the weighing parameters for respective loss components. The optimization objective of the CTB discriminator $D_i$ is given by: 
\begin{equation}
\begin{multlined}
  \mathcal{L}_{D_i} = \frac{1}{2} [ \mathcal{L}_{BCE} (D_i (I_A, I_B), 1) + \\
  \mathcal{L}_{BCE} (D_i (I_A, \overline{I_B}), 0) ]
\end{multlined}
\end{equation}

\noindent
We have used $\lambda_1 = 3$, $\lambda_2 = 1$, $\lambda_3 = 3$, $\lambda_4 = 3$, $\lambda_5 = 10$, $\beta_1 = 5$, $\beta_2 = 1$,  $\beta_3 = 5$ and $\beta_4 = 5$ as the weighing parameters for our experimental setup. The values of these weighing parameters have been empirically estimated through an extensive ablation study.

\section{Experiments}\label{sec:experiments}

\begin{table*}[t]
  \centering
  \caption{Quantitative comparison with SOTA. Results style: \textbf{best}, \underline{second best}. $\uparrow$ higher is better and $\downarrow$ lower is better. Method\textsuperscript{\rm 1} has been pretrained on the same data.}
  \begin{adjustbox}{max width=0.98\textwidth}
  \begin{tabular}{lccccccc}
    \toprule
    Method & pix2pix \cite{isola2017image}  & SRNet\textsuperscript{\rm 1} \cite{wu2019editing}  & SwapText \cite{yang2020swaptext}  & MOSTEL\textsuperscript{\rm 1} \cite{qu2022exploring}  & DiffSTE \cite{ji2023improving}  & \textbf{FASTER}\textsuperscript{\rm 1} & $\Delta$  \\
    \midrule
    PSNR $\uparrow$                           & 12.01  & 18.66  & 19.43 & \underline{20.75}   &  13.40     & \textbf{20.84} & \textcolor{ForestGreen}{\textbf{+0.09}}  \\
    SSIM $\uparrow$                           & 0.349  & 0.614  & 0.652 & \underline{0.707}   &  0.3886    & \textbf{0.790}  & \textcolor{ForestGreen}{\textbf{+0.083}}\\
    FID $\downarrow$                          & 164.24 & 48.16  & \underline{35.62} & 37.55   &  65.06    & \textbf{11.79} & \textcolor{ForestGreen}{\textbf{-23.83}}  \\
    $LPIPS_1$ $\downarrow$               & -      & 0.2076 & -     & \underline{0.1770}  & 0.3958     & \textbf{0.0955} & \textcolor{ForestGreen}{\textbf{-0.0815}} \\
    $LPIPS_2$ $\downarrow$                  & -      & 0.3779 & -     & \underline{0.2895}  & 0.5341     & \textbf{0.1782} & \textcolor{ForestGreen}{\textbf{-0.1113}} \\
    $LPIPS_3$ $\downarrow$                  & -      & 0.2524 & -     & \underline{0.2275}  & 0.4477     & \textbf{0.1227} & \textcolor{ForestGreen}{\textbf{-0.1048}} \\
    Acc (OCR) $\uparrow$     & -      & 14.79  & -     & \underline{20.10}   & 11.90   & \textbf{30.30}& \textcolor{ForestGreen}{\textbf{+10.2}}  \\
    NED (OCR) $\uparrow$ & -      & 0.414  & -     & 0.563   & \textbf{0.714}     & \underline{0.660} & \textcolor{red}{\textbf{-0.054}}  \\ 
    Clip Score $\uparrow$                     & -      & 22.55  & -     & 23.74   & \textbf{25.56}      & \underline{24.00} & \textcolor{red}{\textbf{-1.56}}  \\
    Inf. Time $\downarrow$                & -      & \underline{46.88}  & -     & 76.87   & 12000    & \textbf{31.12} & \textcolor{ForestGreen}{\textbf{-15.76}}  \\
    \bottomrule
  \end{tabular}
  \end{adjustbox}
    \label{tab:comparison}
\end{table*}

\subsection{Datasets}
\noindent
\textbf{Synthetic Data:} For the supervised training of our pipeline, we utilized a dataset of 150,000 labeled images as mentioned in the work of Qu et al.~\cite{qu2022exploring}. For evaluation, we use the Tamper-Syn2k~\cite{qu2022exploring} dataset, which consists of 2,000 paired images specifically designed for evaluation purposes.
We also generated an additional set of 100,000 labeled images using 2,500 fonts. This expanded dataset was employed specifically for training purposes of the proposed method.

\noindent
\textbf{Real Data:} In our research on enhancing natural scene text editing on real scene images, we utilized a dataset consisting of real scene images obtained from the MOSTEL paper~\cite{qu2022exploring}. This dataset is generated by the authors using random cropped images from ICDAR 2013~\cite{karatzas2013icdar},  MLT-2017~\cite{nayef2017icdar2017} MLT-2019~\cite{nayef2019icdar2019} datasets. 

\subsection{Implementation Details}
\noindent
The pre-transformation stage includes resizing the input images to $64 \times 256$. We utilize the 
Adam optimizer~\cite{kingma2014adam} with $\beta_1 = 0.5$ and $\beta_2 = 0.999$, and set the 
learning rate to $1 \times 10^{-3}$ for both the Stages. We train FASTER for a total of 100k iterations using a batch size of 40. Our pipeline is implemented using PyTorch~\cite{imambi2021pytorch} and it is trained on a single NVIDIA 4080 OC GPU. 

\begin{figure} 
  \centering
  \includegraphics[width=0.48\textwidth]{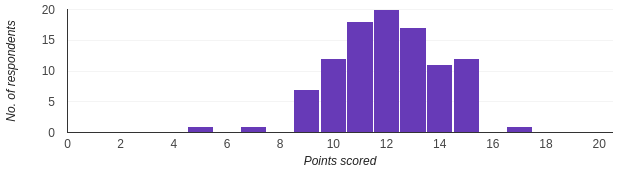} 
  \caption{Human Evaluation Study}
  \label{fig:human}
\end{figure}

\subsection{Evaluation Metrics}
\noindent
To evaluate the effectiveness of FASTER in the STE task, we employ the following commonly used metrics adapted in ~\cite{wu2019editing, yang2020swaptext} to evaluate the generated target style edited image: (1) \texttt{Mean Squared Error (MSE)}, which measures the L-2 distance between the image pair; (2) \texttt{Peak Signal-to-Noise Ratio} (PSNR), which measures the text image quality, and (3) \texttt{Structural Similarity Index Measure} (SSIM)~\cite{wang2004image}, which measures the mean structural similarity between the target and edited sample. We have also used the GAN metrics adopted in~\cite{krishnan2023textstylebrush} for further evaluation for (4) \texttt{Learned Perceptual Image Patch Similarity} (LPIPS)~\cite{zhang2018unreasonable}, which measures the similarities in activation of the paired image patches of Squeezenet~\cite{iandola2016squeezenet}, VGG~\cite{simonyan2014very} and AlexNet~\cite{krizhevsky2012imagenet} denoted as $LPIPS_1$, $LPIPS_2$ and $LPIPS_3$ in Table~\ref{tab:comparison}. (5) \texttt{Frechet Inception Distance (FID)} computes the distance between the feature vectors of the target and edited text images.(6) \texttt{OCR Accuracy and Normalized Edit Distance}~\cite{baek2019wrong}, which calculates the recognition accuracies from the image patches. (7) \texttt{Clip Score} Introduced in ~\cite{hessel2021clipscore}, it computes the semantic similarity between the image and the textual space as in CLIP model~\cite{radford2021learning} and quantifies "compatibility". The reason CLIP Score has been adapted for evaluation is due to its high correlation with human judgment and inspired from ~\cite{chen2023textdiffuser}.

\subsection{Human Evaluation Study} 
\noindent
Human evaluation studies provide crucial insights into how edited and rendered text images are perceived and utilized in real-world scenarios, ensuring that STE models meet end-user expectations. An interesting human evaluation study was conducted with 100 human participants independently asked to mark some samples of images as real (original) or fake (edited). We computed how many images generated by FASTER have been identified as real by the participants. Figure~\ref{fig:human} shows a plot of the points scored by participants based on correctly identified images. An \textbf{overall human misclassification rate of 39.75\%} was recorded which shows the difficulty of the fake identification challenge for the participants created by FASTER. Figure~\ref{fig:vis_1} shows some of the qualitative examples assigned to the user study where participants failed mostly to classify correctly.


\begin{figure*}[t]
    \centering
    \includegraphics[width=.98\textwidth]{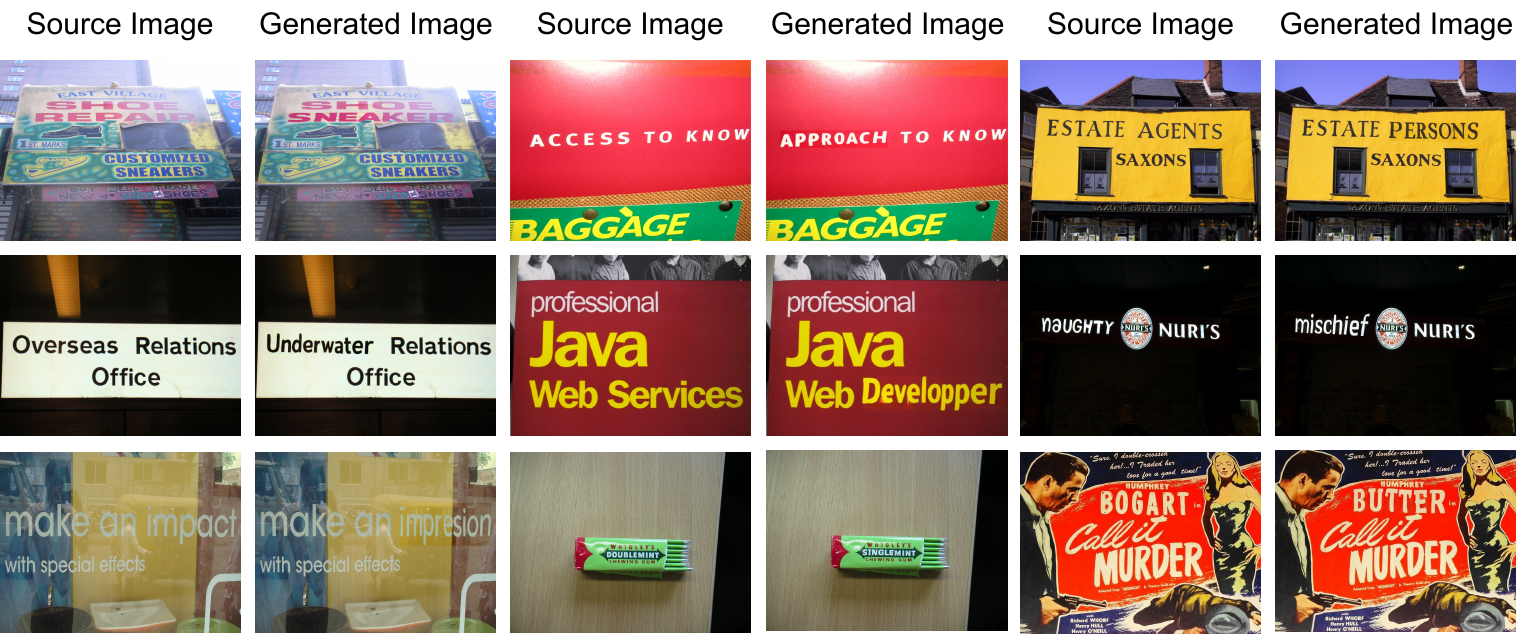}
    \caption{Visual STE results with FASTER on samples from the Real dataset, selected from the human evaluation study with the highest number of incorrect predictions. Please zoom 300\% for better visualization.}
\label{fig:vis_1}
\end{figure*}

\subsection{Comparisons with Existing SOTA}
\noindent
We present a detailed quantitative and qualitative comparison of our proposed method, FASTER, with the existing SOTA methods in STE~\cite{qu2022exploring, ji2023improving, wu2019editing, yang2020swaptext} and the classic pix2pix~\cite{isola2017image} approach. Our model has been trained on the same dataset as used in the MOSTEL paper~\cite{qu2022exploring} and our own synthetically generated data. \\
\noindent
\textbf{Quantitative Analysis:}
 As shown in Table~\ref{tab:comparison}, results demonstrate that our proposed method (FASTER) outperforms all the SOTA approaches in 8 out of the 10 evaluation measures. The key highlights and insights from the results are:\\
 \noindent
 - There has been a \textit{substantial improvement in the perceptual FID Score} with a difference of 23.83 over second-best SwapText. This indicates that our approach generates images with higher similarity to the ground truth and exhibits better perceptual quality.\\
 \noindent
 - The OCR \textit{recognition accuracy} scores have also had a \textit{10\% gain} over second-best MOSTEL, which shows that generated images have excellent legibility and that generated images blend seamlessly with edited content.\\ 
 \noindent
 - We also evaluated with the \textit{inference time} and compared our method with other approaches. FASTER performs editing operations with 15.76 milliseconds less than the second-best SRNet. This justifies the efficiency of the model, especially in sharp contrast with stable diffusion approaches like DiffSTE which takes 12000 milliseconds to operate. \\  
 - The DiffSTE approach shows the best results in terms of ClipScore and Normalized ED (OCR) metrics as they have adapted the Clip model~\cite{radford2021learning} in their training pipeline, which aligns the semantic space between the image and text. \\
\noindent
\textbf{Qualitative Analysis:} As shown in Figure~\ref{fig:comp_3}, we provide visual comparisons of the generated images from different previous methods and FASTER. Notably, the images generated by SRNet demonstrate difficulties in accurately capturing the desired features. While MOSTEL's generated images are of good quality, when compared directly with FASTER, they are not as impressive. It is important to note that although MOSTEL shows better results in some cases compared to FASTER, the visual quality and fidelity of the generated images produced by FASTER, surpass MOSTEL's performance. The most interesting results have been obtained with DiffSTE approaches where the text does not blend well with the image background, which shows the incapability of Stable diffusion models when it comes to rendering precise (fine-grained) image objects like text in scenes. On the contrary, as shown in Figure~\ref{fig:vis_1} the editing results in real data for the FASTER model show that it is highly robust with background distortions. Overall, FASTER showcases superior results both quantitatively and qualitatively when compared to the previous methods considered in the evaluation. 

\begin{figure*}[t]
\centering
\includegraphics[ width=.98\textwidth]{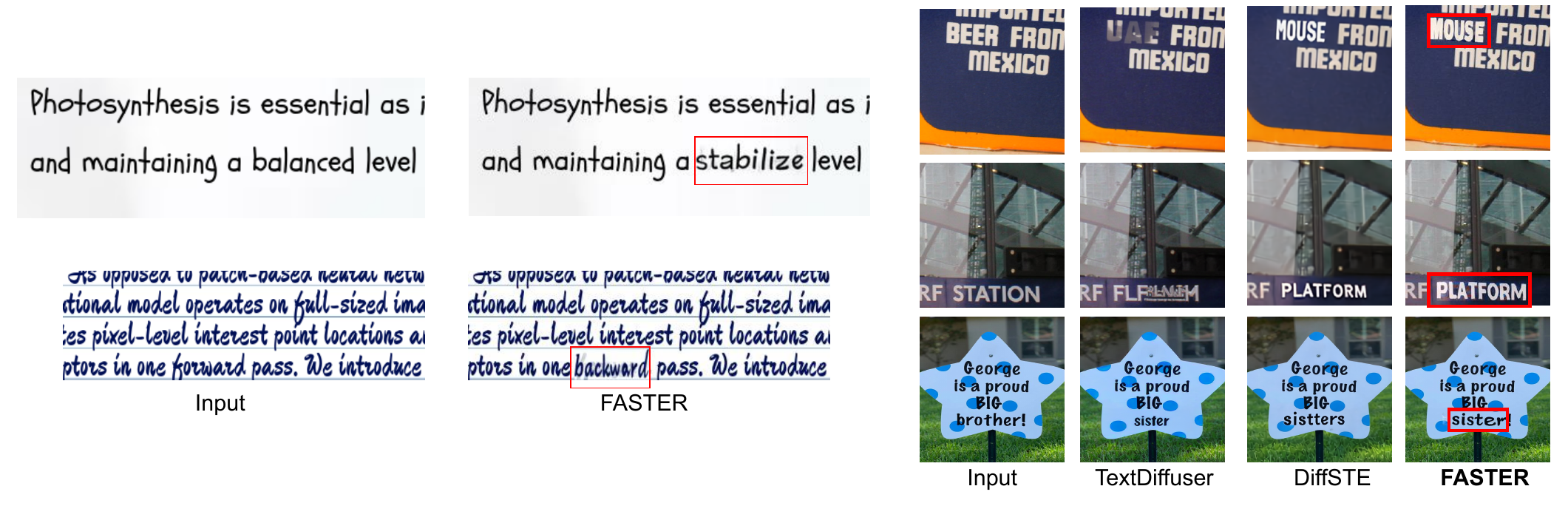}
\caption{\textbf{Left} Fancy Font Text Editing. \textbf{Right} Visual comparison with TextDiffuser and DiffSTE on larger context}
\label{fig:comp_3}
\end{figure*}



\subsection{Ablation Study}
An exhaustive set of experiments has been conducted with the FASTER framework to investigate how the integrated model components have contributed to the overall potential. 

\noindent
\textbf{Effectiveness of Individual Learning Objectives:} To show the individual importance of the learning objectives and how they impact final STE performance, we conducted experiments as shown in Table~\ref{tab:ablation_losses}. It throws some quantitative insights in terms of different metrics. Adding OCR Perceptual loss~\cite{rodriguez2023ocr} term to the training objective helps minimise the distance between the target and edited text samples in the feature space. Table~\ref{tab:ablation_losses} show that \textit{$\mathcal{L}_{{OCR}_{per}}$  has the highest impact on overall performance metrics} (especially FID, PSNR and MSE scores). While the VGG perceptual loss also leaves a reasonably good individual impact, it is interesting to note that it decreases the content evaluation metrics (OCR Accuracy and NED). This can be attributed to the fact that it adds noise to the already high-performant OCR perceptual loss. 

\begin{table}[t]

\centering
\caption{Effectiveness of individual losses on FASTER. Results style: \textbf{best}, \underline{second best}. Results style: \textbf{best}, \underline{second best}.}
\begin{adjustbox}{max width=0.48\textwidth}
\begin{tabular}{lcccccccccc}
\toprule
Method & MSE $\downarrow$ & PSNR $\uparrow$ & SSIM $\uparrow$ & FID $\downarrow$ & $LPIPS_1$ $\downarrow$ & $LPIPS_2$ $\downarrow$ & $LPIPS_3$ $\downarrow$ & Acc $\uparrow$ &  NED $\uparrow$ & Clip Score $\uparrow$ \\
\midrule
w/o $\mathcal{L}_{GAN}$       & 0.0200 & 18.41 & 0.6889 & 23.56 & 0.1332 & 0.2431 & 0.1651 & 30.65 & 0.654 & 23.89\\
w/o $\mathcal{L}_{L_2}^{G_m}$ & 0.0200 & 18.40 & 0.6880 & \underline{19.58} &0.1327 & 0.2436 & 0.1657 & 29.40 & 0.652 & 23.90\\
w/o $\mathcal{L}_{SSIM}^{G_m}$       & \underline{0.0198} & \underline{18.47} & \underline{0.6907} & 23.47 & \underline{0.1309} & \underline{0.2409} & \underline{0.1633} & 30.75 & 0.655 & 23.90  \\
w/o $\mathcal{L}_{P_{\rho}}$  & 0.0201 & 18.40 &  0.6878 & 23.60 & 0.1327 & 0.2441 & 0.1653 & \textbf{31.15} & \textbf{0.662 }& \underline{23.78}\\
w/o $\mathcal{L}_{{OCR}_{per}}$ & 0.0224 & 18.00 & 0.6775  & 24.59 &0.1441 & 0.2598 & 0.1782 &  19.20 & 0.569 & 23.89\\
All              & \textbf{0.0127} & \textbf{20.85}& \textbf{0.7905} & \textbf{11.79}  & \textbf{0.0955} & \textbf{0.1782} & \textbf{0.1227} & \underline{30.30}& \underline{0.660} & \textbf{24.00} \\ 
\bottomrule
\end{tabular}
\end{adjustbox}
\label{tab:ablation_losses}
\end{table}

\noindent
\textbf{Effectiveness of Data Mixing:} In this study, we utilized two datasets: the MOSTEL dataset for initial training and a mixed dataset combining MOSTEL and SRNET data. Results from Table \ref{tab:data} demonstrate that the mixed data model performs better than the single dataset model. 

\noindent
\textbf{How effective is the Cascaded Self-Attention in USBs?} We investigated the influence of different attention mechanisms on image upsampling in FASTER. Initially, we evaluated the performance of isolated sigmoid attention by incorporating four sigmoid attention blocks in the USBs. Subsequently, we examined the integration of four self-attention blocks within the USB. Finally, combining two sigmoid and two self-attention blocks yielded superior metrics compared to previous experiments, as shown in Table \ref{tab:attention} using the dataset in ~\cite{qu2022exploring}.

\begin{table}[t]
    \begin{minipage}{.48\linewidth}
    \caption{Effectiveness of Data Mixing.}
        \begin{adjustbox}{max width=0.9\textwidth}
  \begin{tabular}{lrrr}
    \toprule
    Dataset        & MSE $\downarrow$ & PSNR $\uparrow$  & SSIM $\uparrow$ \\
    \midrule
    MOSTEL \cite{qu2022exploring} & 0.0171           & 19.04          & 0.707           \\
    Mixed          & \textbf{0.0162}  & \textbf{19.19} & \textbf{0.718}   \\
    \bottomrule
  \end{tabular}  
\end{adjustbox}
 
  \label{tab:data}
    \end{minipage}%
    \begin{minipage}{.48\linewidth}
      \centering
      \caption{Effect of different USBs.}
  \begin{adjustbox}{max width=0.9\textwidth}
  \begin{tabular}{ccccccc}
    \toprule
    \multicolumn{4}{c}{Attentions in USB} \\
    \cmidrule(r){1-4}
    1st & 2nd &  3rd &  4th & MSE $\downarrow$ &  PSNR $\uparrow$ &  SSIM $\uparrow$ \\
    \midrule
    $\sigma$ & $\sigma $ & $\sigma $ & $\sigma $ & 0.0216 & 18.22 & 0.690 \\
    SA      &      SA &      SA &      SA & 0.0213 & 18.00 & 0.672 \\
    SA      &      SA & $\sigma $ & $\sigma $ & \textbf{0.0171} & \textbf{19.04} & \textbf{0.707} \\
    \bottomrule
  \end{tabular}
  \end{adjustbox}
   
    \label{tab:attention}
    \end{minipage} 
\end{table}
\noindent
\textbf{Concatenation Strategies and Input Combination:} Initially, our proposed pipeline was guided by concatenating the input mask $m_A$ with a fixed mask $m_F$. Then, we experimented to assess the outcome when only the fixed mask $m_F$ was utilized without concatenation. Additionally, we evaluated the effectiveness of producing results from both binary source mask and image $I_A$ as shown in Table~\ref{tab:inputs}.

\begin{table}[t]
    \begin{minipage}{.48\linewidth}
    \caption{Effectiveness of Inputs.}
        \begin{adjustbox}{max width=0.9\textwidth}
  \begin{tabular}{llrrr}
    \toprule
    \multicolumn{2}{l}{Input Combination} \\
    \cmidrule(r){1-2}
     Input Type & $m_a$ Concat & MSE $\downarrow$ &  PSNR $\uparrow$ &  SSIM $\uparrow$ \\
    \midrule
    Mask    & Concat     & 0.0162  & 19.19 & 0.718  \\
    Mask    & w/o Concat & 0.0178 & 18.86 & 0.701 \\ 
    Image   & Concat     & \textbf{0.0135} & \textbf{20.20} & \textbf{0.776} \\
    Image   & w/o Concat & 0.0156 & 19.55 & 0.759 \\ 
    \bottomrule
  \end{tabular}
\end{adjustbox}
  \label{tab:inputs}
    \end{minipage}%
    \begin{minipage}{.48\linewidth}
      \centering
      \caption{Effectiveness of different Fonts.}
  \begin{adjustbox}{max width=0.9\textwidth}
  \centering
  \begin{tabular}{lccc}
    \toprule
    Fonts & MSE $\downarrow$ & PSNR $\uparrow$ & SSIM $\uparrow$ \\
    \midrule
    Sans Serif      & 0.0270 & 17.16 & 0.6432 \\
    Times New Roman & 0.0273 & 17.11 & 0.6372 \\
    Arial           & \textbf{0.0264} & \textbf{17.27} & \textbf{0.6493} \\
    \bottomrule
  \end{tabular}
  \end{adjustbox}
  \label{tab:font_ablation}
    \end{minipage} 
\end{table}

\noindent
\textbf{Exploring Font Style and Size Variations:} As shown in Table~\ref{tab:font_ablation} and Table~\ref{tab:font_size_ablation}, we conducted an ablation study to investigate the impact of different font types and sizes on FASTER's performance. The results indicate that varying fonts (Arial, Times New Roman, Sans Serif) and sizes (25 and 28) have minimal impact. Our synthetic training dataset includes diverse fonts and sizes, similar to a standard OCR dataset. These findings highlight FASTER's robustness in handling diverse font styles and sizes.

\begin{table}[t]
    \begin{minipage}{.45\linewidth}
    \centering
\caption{Effectiveness of font sizes.}
        \begin{adjustbox}{max width=0.9\textwidth}

  \begin{tabular}{lccc}
    \toprule
    Method & MSE $\downarrow$ & PSNR $\uparrow$ & SSIM $\uparrow$ \\
    \midrule
    Arial font size 28 & 0.0283 & 16.90 & 0.6353 \\
    Arial font size 25 & \textbf{0.0264} & \textbf{17.27} & \textbf{0.6493} \\
    \bottomrule
  \end{tabular}

\end{adjustbox}
  
  \label{tab:font_size_ablation}
    \end{minipage}%
    \begin{minipage}{.48\linewidth}
      \centering
      \caption{Effectiveness of two stages.}
  \begin{adjustbox}{max width=0.9\textwidth}
\centering
  \begin{tabular}{lccc}
    \toprule
    Method & MSE $\downarrow$ & PSNR $\uparrow$ & SSIM $\uparrow$ \\
    \midrule
    Jointly       & 0.0311 & 17.14 & 0.6773 \\
    Independently & \textbf{0.0135} & \textbf{20.20} & \textbf{0.7760} \\
    \bottomrule
  \end{tabular}
  \end{adjustbox}

\label{tab:stages_ablation}
    \end{minipage} 
\end{table}

\noindent
\textbf{FASTER benefits more from independent stage training compared to joint training.} FASTER consist of two stages (STB and CTB) in the overall pipeline. As demonstrated in Table~\ref{tab:stages_ablation},  The "Two stages Independently" strategy showcases noteworthy outcomes with an impressive MSE reduction to 0.013, a notable PSNR increase to 20.20, and a significant SSIM enhancement to 0.77. By training style and content blocks independently, each block can specialize in its respective task, leading to more effective learning and adaptation to the characteristics of the input data compared to less favourable "Two stages jointly". 



\section{Conclusion and Future Scope}\label{sec:conclusion}

\noindent

Current STE methods often prioritize image quality over practical usability. To address this, we introduce FASTER, a two-stage GAN-based framework that seamlessly edits and renders text while preserving its natural appearance. The cascaded self-attention module enhances processing speed and editing capabilities. FASTER is robust across various text attributes (font type, size, alignment) and domains (scene text, documents, handwriting), making it ideal for real-world applications.


The main limitation of our method is its reliance on mask guidance maps to identify editing regions. While this helps eliminate background distractions and focus on text regions, it struggles with complex or irregular text layouts. The guidance maps assume well-defined masks, which is problematic for diverse text forms like complex handwriting, irregular, or overlapping text (see Supplementary). Generating accurate masks in these cases can be challenging and may lead to errors.


{\small
\bibliographystyle{ieee_fullname}
\bibliography{egbib}
}

\newpage

\appendix
\onecolumn
\section*{\noindent \huge Supplementary Material for \texttt{FASTER}}
\begin{figure*}[b]
\centering
\includegraphics[ width=.98\textwidth]{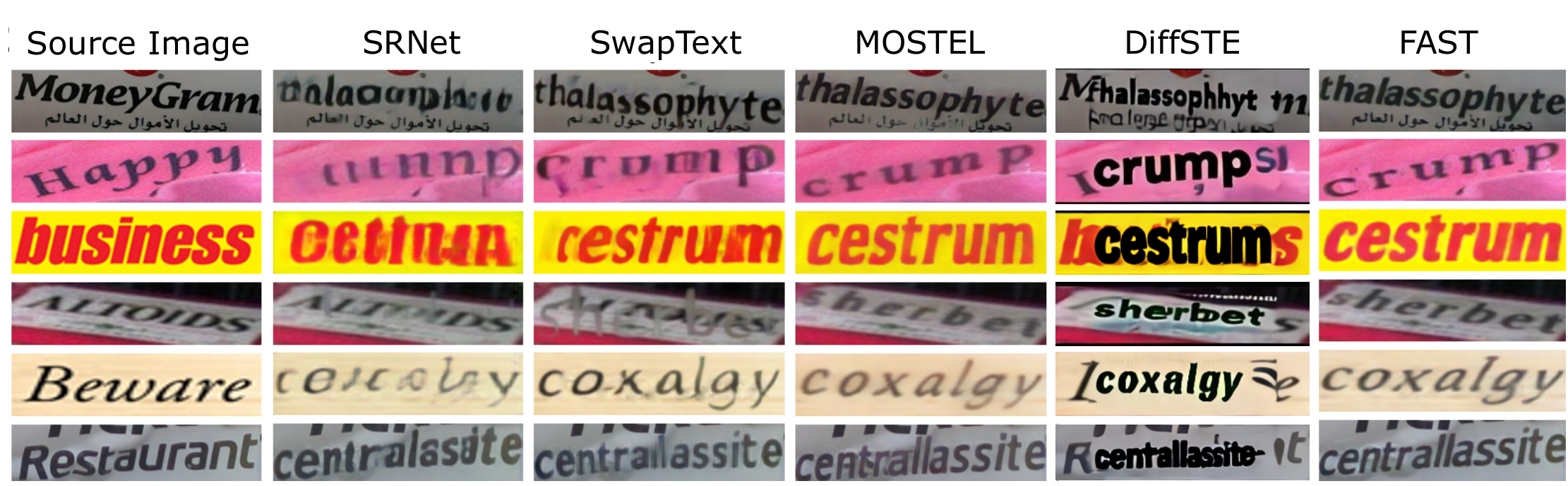}
\caption{Visual comparison of FASTER with current SOTA methods on real scenes.}
\label{fig:comp_3}
\end{figure*}


\paragraph{Synthetic Data:} For the supervised training of our pipeline, we utilized a dataset of 150,000 labeled images as mentioned in the work of Qu et al.~\cite{qu2022exploring}. For evaluation, we use the Tamper-Syn2k~\cite{qu2022exploring} dataset, which consists of 2,000 paired images specifically designed for evaluation purposes. These paired images were created by rendering different texts with consistent styles, including font, size, color, spatial transformation, and background image. To generate these paired images, a collection of 300 fonts and 12,000 background images are used. These background images were subjected to random rotation, curve, and perspective transformations to introduce diversity. It is worth noting that we also generated an additional set of 100,000 labeled images using 2,500 fonts. This expanded dataset was employed specifically for training purposes of the proposed method.

\paragraph{Real Data:} In our research on enhancing natural screen text editing on real scene images, we utilized a dataset consisting of real scene images obtained from the MOSTEL~\cite{qu2022exploring}. This dataset is generated by the authors using random cropped images from ICDAR 2013~\cite{karatzas2013icdar},  MLT-2017~\cite{nayef2017icdar2017} MLT-2019~\cite{nayef2019icdar2019} datasets.

\section{Implementation Details}

The implementation process involved training the U-Net backbone using input images resized to $64 \times 256$, utilizing a synthetic dataset as outlined in our main submitted draft. \textit{We employ the U-Net to approximate the binary mask of the input image}. The input images are the same as those taken by the CTB Block of our model, and corresponding binary masks are provided in the dataset as ground truth. Our approach utilized the Adam optimizer~\cite{kingma2014adam} with a learning rate set at 0.001. The U-Net underwent training for a total of 20 epochs, with each batch containing a single iteration. The complete pipeline was implemented using PyTorch~\cite{imambi2021pytorch} and trained on a single NVIDIA 4080 OC GPU.
\begin{figure*}[b]
\centering
\begin{adjustbox}{max width=\textwidth}
\begin{tabular}{cccccc} 
Source Image & Generated Image &  Source Image & Generated Image & Source Image & Generated Image \\\\
\includegraphics[height=0.75cm, width= 3cm]{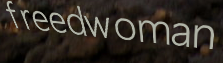} &
\includegraphics[height=0.75cm, width= 3cm]{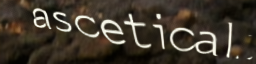} &
\includegraphics[height=0.75cm, width= 3cm]{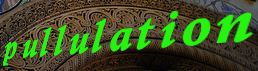} &
\includegraphics[height=0.75cm, width= 3cm]{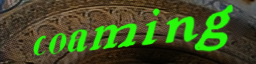} &
\includegraphics[height=0.75cm, width= 3cm]{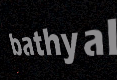} &
\includegraphics[height=0.75cm, width= 3cm]{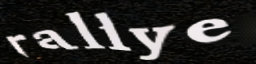}\\
\end{tabular}
\end{adjustbox}
\caption{Some failure cases of our model. }
\label{fig:vis_3}

\end{figure*}
\begin{figure*}[h]
\centering
\begin{adjustbox}{max width=\textwidth}
\begin{tabular}{ccccccccc} 
Source Image & Target Reference &  SRNet \cite{wu2019editing} & MOSTEL \cite{qu2022exploring} & DiffSTE~\cite{ji2023improving} & \textbf{FASTER} \\\\
\includegraphics[height=0.75cm, width= 3cm]{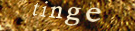} &
\includegraphics[height=0.75cm, width= 3cm]{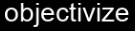} &
\includegraphics[height=0.75cm, width= 3cm]{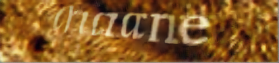} &
\includegraphics[height=0.75cm, width= 3cm]{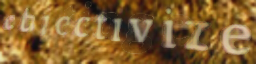} &
\includegraphics[height=0.75cm, width= 3cm]{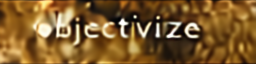} &
\includegraphics[height=0.75cm, width= 3cm]{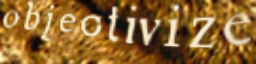} \\

\includegraphics[height=0.75cm, width= 3cm]{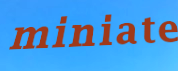} &
\includegraphics[height=0.75cm, width= 3cm]{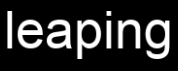} &
\includegraphics[height=0.75cm, width= 3cm]{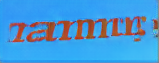} &
\includegraphics[height=0.75cm, width= 3cm]{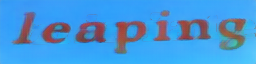} &
\includegraphics[height=0.75cm, width= 3cm]{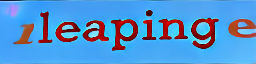} &
\includegraphics[height=0.75cm, width= 3cm]{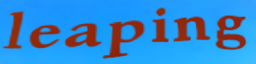} \\

\includegraphics[height=0.75cm, width= 3cm]{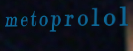} &
\includegraphics[height=0.75cm, width= 3cm]{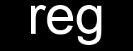} &
\includegraphics[height=0.75cm, width= 3cm]{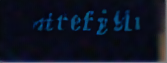} &
\includegraphics[height=0.75cm, width= 3cm]{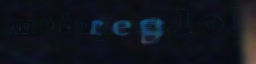} &
\includegraphics[height=0.75cm, width= 3cm]{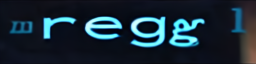} &
\includegraphics[height=0.75cm, width= 3cm]{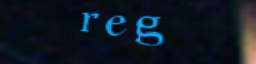} \\

\includegraphics[height=0.75cm, width= 3cm]{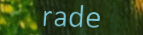} &
\includegraphics[height=0.75cm, width= 3cm]{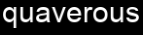} &
\includegraphics[height=0.75cm, width= 3cm]{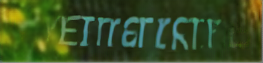} &
\includegraphics[height=0.75cm, width= 3cm]{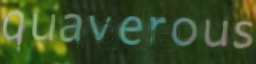} &
\includegraphics[height=0.75cm, width= 3cm]{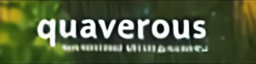} &
\includegraphics[height=0.75cm, width= 3cm]{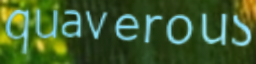} \\

\includegraphics[height=0.75cm, width= 3cm]{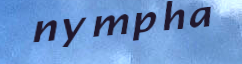} &
\includegraphics[height=0.75cm, width= 3cm]{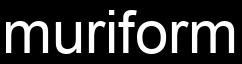} &
\includegraphics[height=0.75cm, width= 3cm]{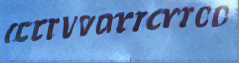} &
\includegraphics[height=0.75cm, width= 3cm]{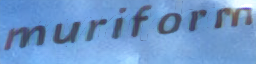} &
\includegraphics[height=0.75cm, width= 3cm]{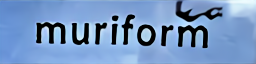} &
\includegraphics[height=0.75cm, width= 3cm]{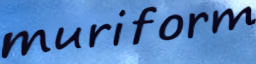} \\

\includegraphics[height=0.75cm, width= 3cm]{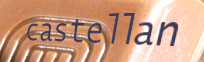} &
\includegraphics[height=0.75cm, width= 3cm]{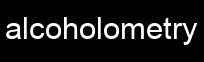} &
\includegraphics[height=0.75cm, width= 3cm]{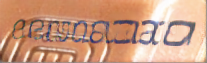} &
\includegraphics[height=0.75cm, width= 3cm]{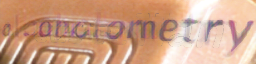} &
\includegraphics[height=0.75cm, width= 3cm]{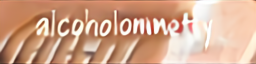} &
\includegraphics[height=0.75cm, width= 3cm]{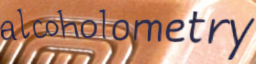} \\

\includegraphics[height=0.75cm, width= 3cm]{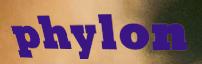} &
\includegraphics[height=0.75cm, width= 3cm]{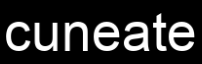} &
\includegraphics[height=0.75cm, width= 3cm]{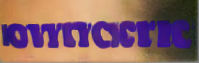} &
\includegraphics[height=0.75cm, width= 3cm]{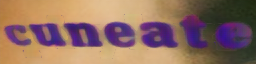} &
\includegraphics[height=0.75cm, width= 3cm]{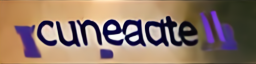} &
\includegraphics[height=0.75cm, width= 3cm]{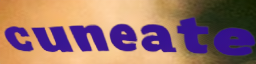} \\

\includegraphics[height=0.75cm, width= 3cm]{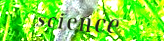} &
\includegraphics[height=0.75cm, width= 3cm]{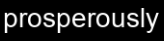} &
\includegraphics[height=0.75cm, width= 3cm]{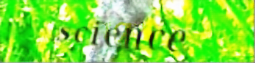} &
\includegraphics[height=0.75cm, width= 3cm]{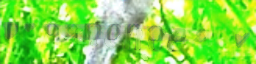} &
\includegraphics[height=0.75cm, width= 3cm]{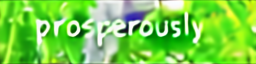} &
\includegraphics[height=0.75cm, width= 3cm]{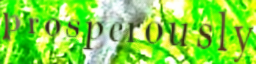} \\
\end{tabular}
\end{adjustbox}
\caption{Qualitative comparison of FASTER with SOTA on Synthetic Data samples}
\label{fig:compare}
\end{figure*}

\begin{figure*}[ht]
\centering
\begin{adjustbox}{max width=\textwidth}
\begin{tabular}{cccccc} 
\includegraphics[height=0.75cm, width= 3cm]{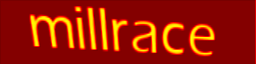} &
\includegraphics[height=0.75cm, width= 3cm]{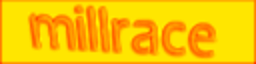} &
\includegraphics[height=0.75cm, width= 3cm]{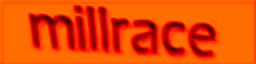} &
\includegraphics[height=0.75cm, width= 3cm]{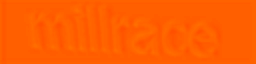} &
\includegraphics[height=0.75cm, width= 3cm]{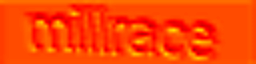} &
\includegraphics[height=0.75cm, width= 3cm]{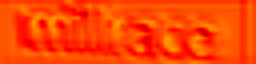} \\
\includegraphics[height=0.75cm, width= 3cm]{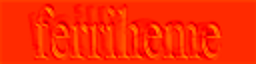} &
\includegraphics[height=0.75cm, width= 3cm]{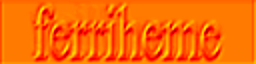} &
\includegraphics[height=0.75cm, width= 3cm]{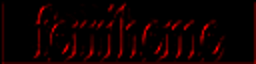} &
\includegraphics[height=0.75cm, width= 3cm]{figures/attention_map/module.in2_down1.3.layers.2.png} &
\includegraphics[height=0.75cm, width= 3cm]{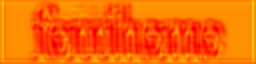} &
\includegraphics[height=0.75cm, width= 3cm]{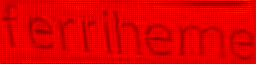} \\
\includegraphics[height=0.75cm, width= 3cm]{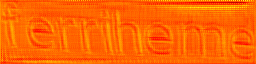} &
\includegraphics[height=0.75cm, width= 3cm]{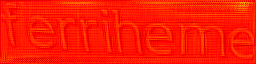} &
\includegraphics[height=0.75cm, width= 3cm]{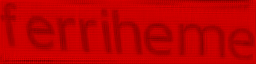} &
\includegraphics[height=0.75cm, width= 3cm]{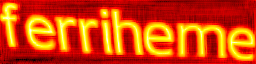} &
\includegraphics[height=0.75cm, width= 3cm]{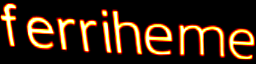} &
\includegraphics[height=0.75cm, width= 3cm]{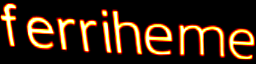} \\
\end{tabular}
\end{adjustbox}
\caption{\textbf{How FASTER Works:} Attention visualization of different phases of image editing using FASTER on a source image with "millrace" and target image with "ferriheme" (from top left to bottom right in order)}
\label{fig:vis_attention}
\end{figure*}

\begin{figure*}[ht]
\centering
\begin{adjustbox}{max width=\textwidth}
\begin{tabular}{cccccc} 
Source Image & Generated Image &  Source Image & Generated Image & Source Image & Generated Image \\\\
\includegraphics[height=0.75cm, width= 3cm]{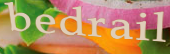} &
\includegraphics[height=0.75cm, width= 3cm]{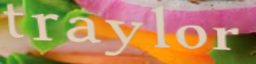} &
\includegraphics[height=0.75cm, width= 3cm]{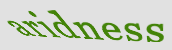} &
\includegraphics[height=0.75cm, width= 3cm]{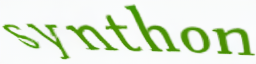} &
\includegraphics[height=0.75cm, width= 3cm]{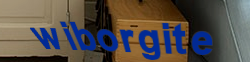} &
\includegraphics[height=0.75cm, width= 3cm]{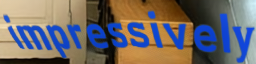} \\
\includegraphics[height=0.75cm, width= 3cm]{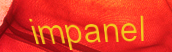} &
\includegraphics[height=0.75cm, width= 3cm]{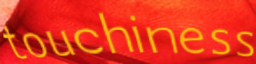} &
\includegraphics[height=0.75cm, width= 3cm]{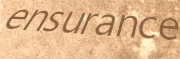} &
\includegraphics[height=0.75cm, width= 3cm]{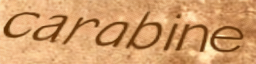} &
\includegraphics[height=0.75cm, width= 3cm]{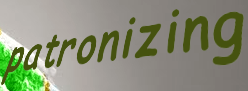} &
\includegraphics[height=0.75cm, width= 3cm]{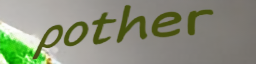} \\
\end{tabular}
\end{adjustbox}
\caption{Qualitative image editing using FASTER on some arbitrary-shaped text examples from Synthetic dataset}
\label{fig:vis_2}
\end{figure*}

\begin{figure*}[ht]
\centering
\begin{adjustbox}{max width=\textwidth}
\begin{tabular}{cccccc} 
Source Image & Generated Image &  Source Image & Generated Image & Source Image & Generated Image \\\\
\includegraphics[height=0.75cm, width= 3cm]{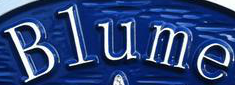} &
\includegraphics[height=0.75cm, width= 3cm]{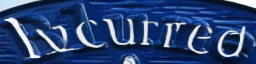} &
\includegraphics[height=0.75cm, width= 3cm]{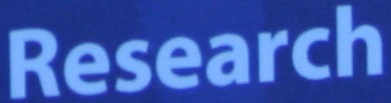} &
\includegraphics[height=0.75cm, width= 3cm]{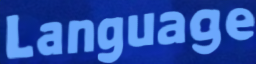} &
\includegraphics[height=0.75cm, width= 3cm]{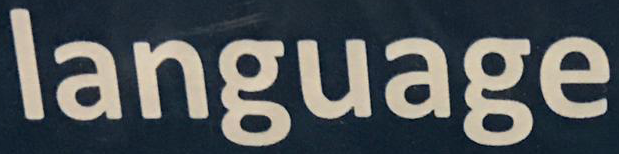} &
\includegraphics[height=0.75cm, width= 3cm]{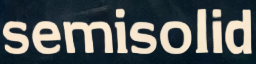} \\
\includegraphics[height=0.75cm, width= 3cm]{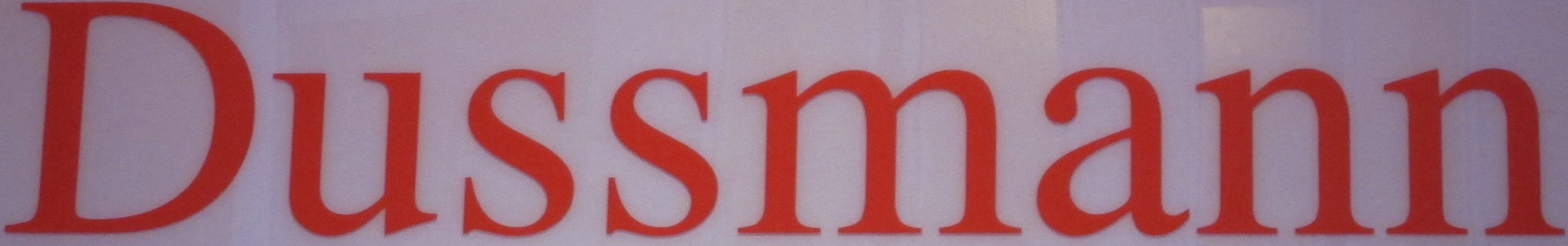} &
\includegraphics[height=0.75cm, width= 3cm]{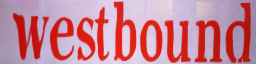} &
\includegraphics[height=0.75cm, width= 3cm]{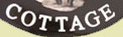} &
\includegraphics[height=0.75cm, width= 3cm]{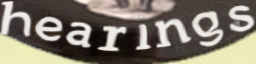} &
\includegraphics[height=0.75cm, width= 3cm]{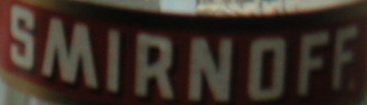} &
\includegraphics[height=0.75cm, width= 3cm]{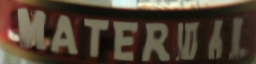} \\
\includegraphics[height=0.75cm, width= 3cm]{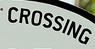} &
\includegraphics[height=0.75cm, width= 3cm]{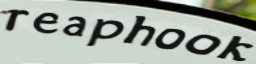} &
\includegraphics[height=0.75cm, width= 3cm]{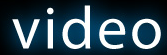} &
\includegraphics[height=0.75cm, width= 3cm]{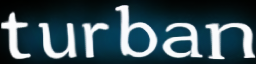} &
\includegraphics[height=0.75cm, width= 3cm]{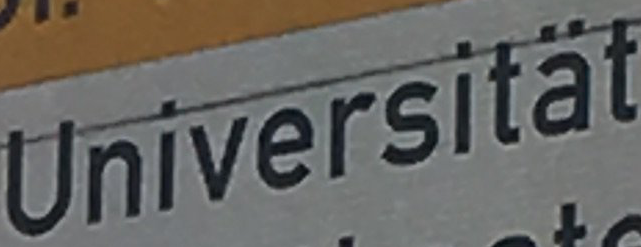} &
\includegraphics[height=0.75cm, width= 3cm]{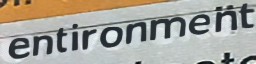} \\
\end{tabular}
\end{adjustbox}
\caption{Qualitative image editing examples using FASTER on the Real dataset. }
\label{fig:vis_1}
\end{figure*}

\begin{figure*}[t]
  \begin{subfigure}{.5\textwidth}
  \centering
    \includegraphics[width=1\linewidth]{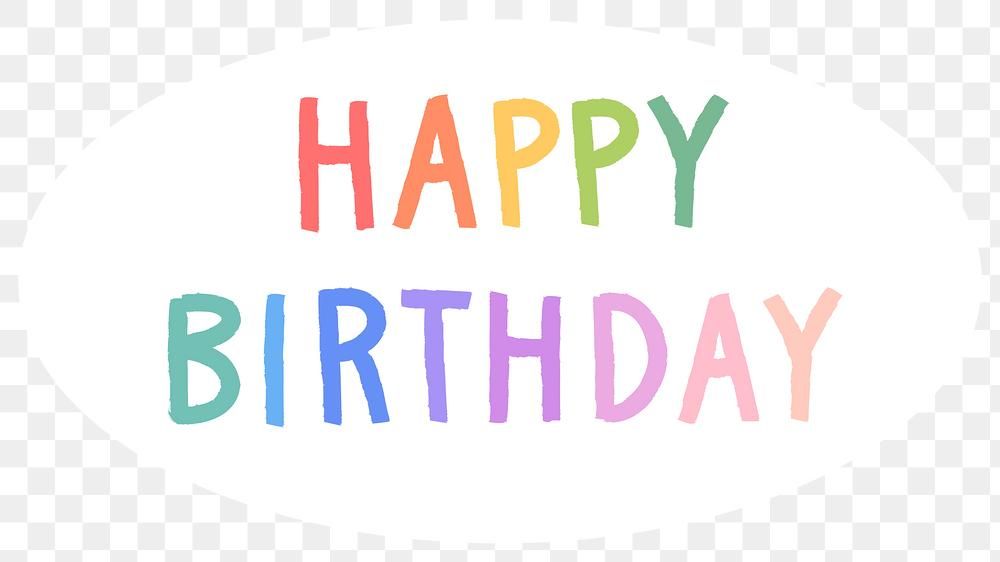}
  \end{subfigure}%
  \begin{subfigure}{.5\textwidth}
  \centering
    \includegraphics[width=1\linewidth]{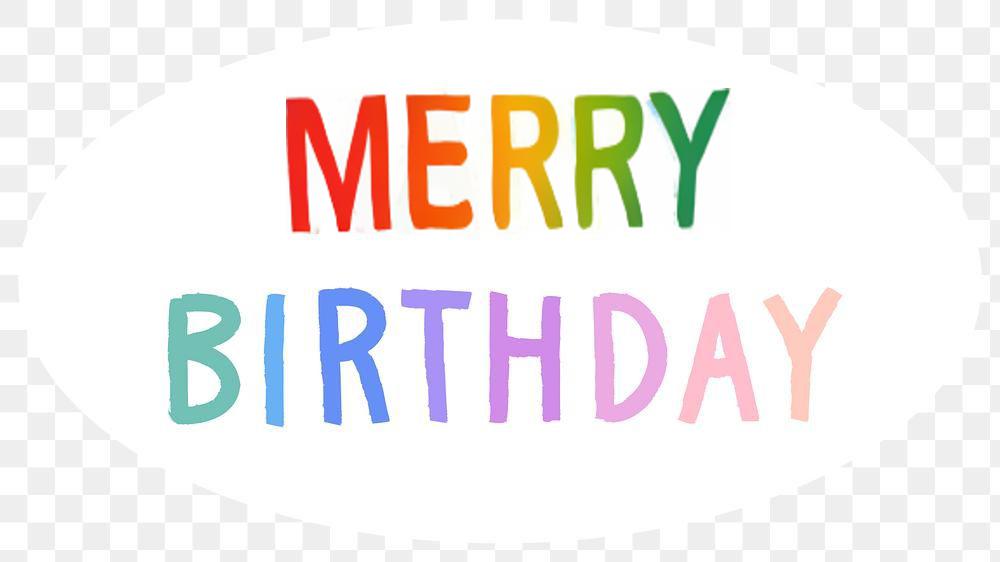}
  \end{subfigure}
  \begin{subfigure}{.5\textwidth}
  \centering
    \includegraphics[width=1\linewidth, height = 8cm]{figures/supply_img/0000176.jpg}
  \end{subfigure}%
  \begin{subfigure}{.5\textwidth}
  \centering
    \includegraphics[width=1\linewidth, height = 8cm]{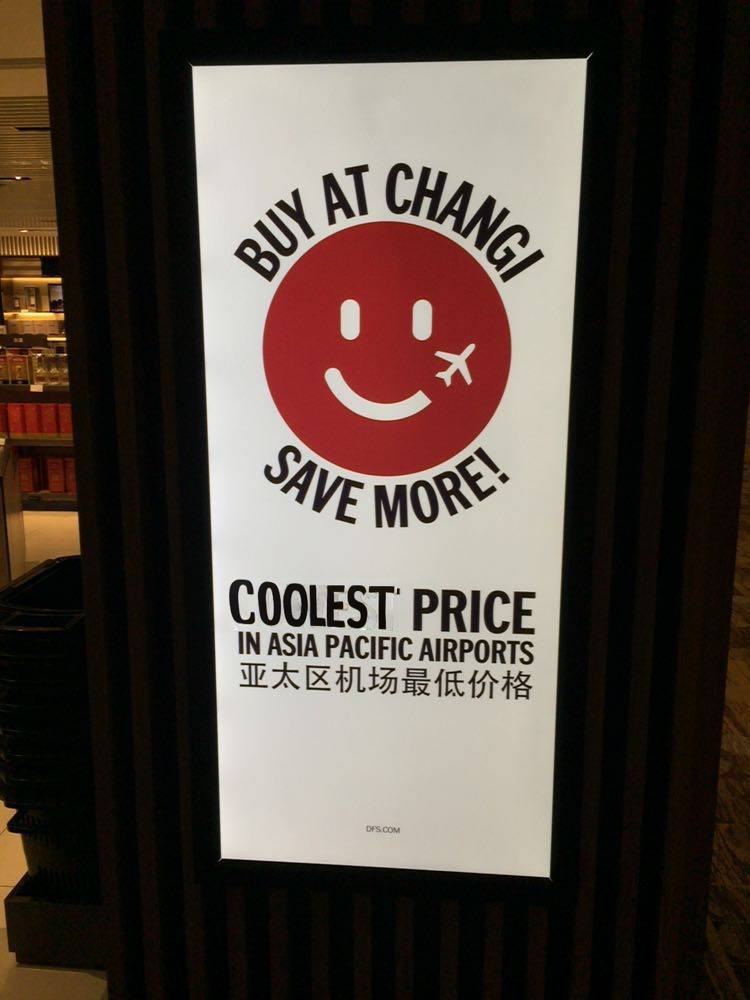}
  \end{subfigure}
  \begin{subfigure}{.5\textwidth}
  \centering
    \includegraphics[width=1\linewidth]{figures/supply_img/0000233.jpg}
  \end{subfigure}%
  \begin{subfigure}{.5\textwidth}
  \centering
    \includegraphics[width=1\linewidth]{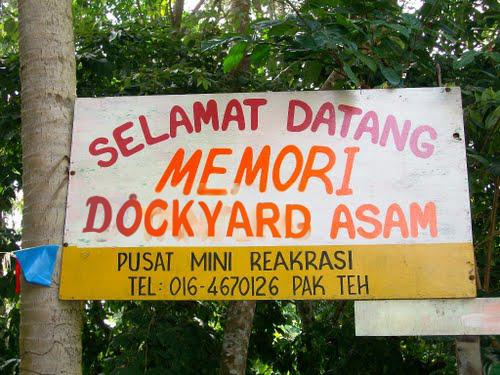}
  \end{subfigure}
  \caption{Qualitative evaluation of editing on real images from Internet}
  \label{fig: first_case}
\end{figure*}


\section{Further Analysis and Discussions}

\paragraph{\textbf{Not Just Another STE Method}}

Please note that our method is not just
"another method" that provides better scene text editing. Our work aims to develop a novel \textit{font-agnostic method} that simultaneously generates text in arbitrary styles and locations while preserving a
natural and realistic appearance through a simple combination of mask generation and text style transfer as shown by \textbf{attention map visualization} in Figure~\ref{fig:vis_attention}. To the best of our
knowledge, this is the first work of its kind. Moreover, our approach differs from the existing methods as they directly modify all image pixels. Instead, the proposed method has introduced a filtering mechanism
to remove background distractions, allowing the network to focus solely on the text regions where
editing is required as shown in some qualitative examples in Figure~\ref{fig:compare} for synthetic and Figure~\ref{fig:vis_1} for real images. Additionally, our results elegantly spotlight our method's prowess across varied scene text styles. We also see in Figure~\ref{fig:vis_2} some examples of images containing arbitrarily shaped text where the model still can perform well. 

\paragraph{\textbf{FASTER attains the best results (in terms of average inference time) when compared to existing SOTA STE models.}}

In Table 1, we have thoughtfully presented the average time taken by each of the methods under consideration. This table serves as a valuable reference point for assessing the efficiency of the different approaches. Specifically, it is noteworthy that that "FASTER" stands out in terms of its computational speed during inference. The model \textit{clocks an impressive speed of 31.12 milliseconds for generating a single image}. This efficiency is not only remarkable in isolation but also positions "FASTER" as a significantly swifter option compared to the previously employed methodologies specially \textit{compared to stable diffusion-based~\cite{ji2023improving} approaches which take almost 400 times more time for single image inference}. This observation underscores the strides made in optimizing the computational efficiency of our approach, highlighting its potential to expedite the text image editing operation effectively.



\paragraph{\textbf{What makes FASTER establish font-agnostic text editing?}}

As demonstrated in Table 6 and Table 7, we examine how different font types and sizes impact our outcomes. This is an ablation study, not a comparison study with SOTA methods and hence incremental accuracy gain is not applicable here. The results from Table 6, where \textit{we explore Arial, Times New Roman, and Sans Serif fonts, indicate minimal influence on our method's performance}. Similarly, Table 7, maintaining \textit{different sizes (25 and 28) of fixed font, also showcases negligible effects.} Importantly, it's worth noting that our training synthetic dataset encompasses diverse fonts and sizes, being a standard OCR dataset. These collective findings underscore our method's font-agnostic nature, where it consistently performs across font variations. It can be noted that if we change different fonts and font sizes, the accuracy of our method does not change much. Hence, our method is robust in different fonts and font sizes.

\section{Visualizing Real-World Scene Editing}

This section highlights some of the key observations, qualitative analysis and case studies to demonstrate the potential of FASTER in real-world scene editing applications.

\paragraph{\textbf{Advertising and Marketing:}} 
STE allows marketers to update product details, prices, and promotional messages in real-world advertisements, posters, and billboards efficiently and effectively. As shown in Figure~\ref{fig: first_case}, the second example shows an advertisement board, which preserves the background style and structural properties while editing text seamlessly between "LOWEST" and "COOLEST". Also, in third example use-case as shown we see the content edit between "DTANJUNG" and generated "DOCKYARD" where the content is in a different language (Malaysian/Indonesian).  

\paragraph{\textbf{Retail and E-commerce:}}

\begin{figure*}[ht]
  \begin{subfigure}{.5\textwidth}
  \centering
    \includegraphics[width=1\linewidth]{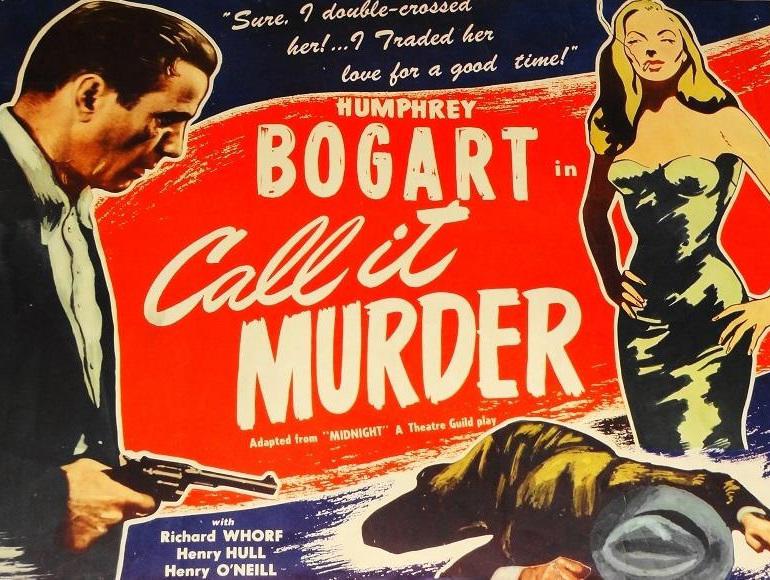}
  \end{subfigure}%
  \begin{subfigure}{.5\textwidth}
  \centering
    \includegraphics[width=1\linewidth]{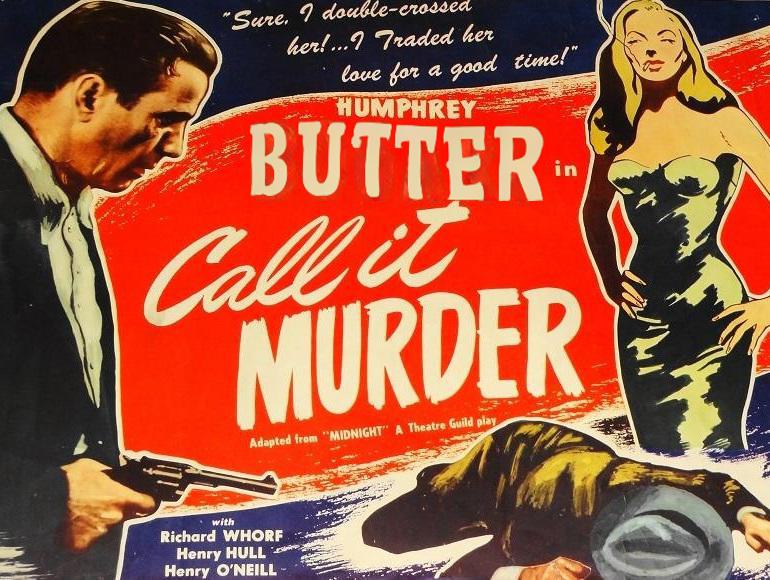}
  \end{subfigure}
  \begin{subfigure}{.5\textwidth}
  \centering
    \includegraphics[width=1\linewidth]{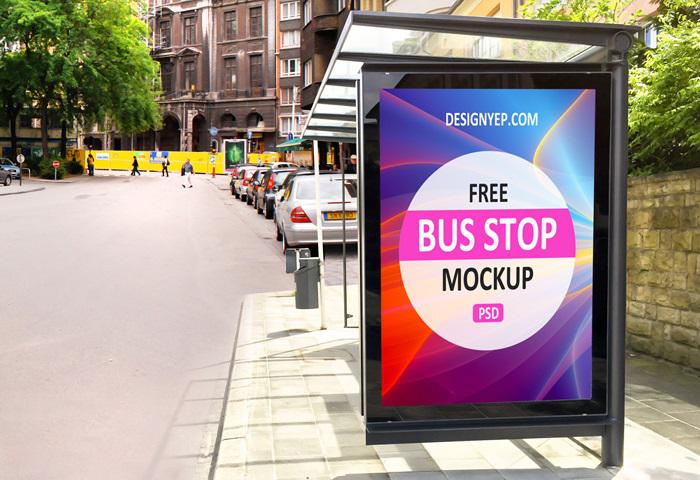}
  \end{subfigure}%
  \begin{subfigure}{.5\textwidth}
  \centering
    \includegraphics[width=1\linewidth]{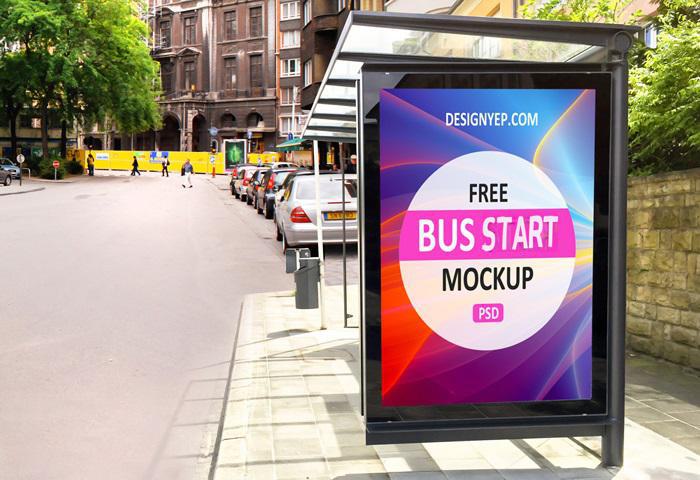}
  \end{subfigure}
  \begin{subfigure}{.5\textwidth}
  \centering
    \includegraphics[width=1\linewidth]{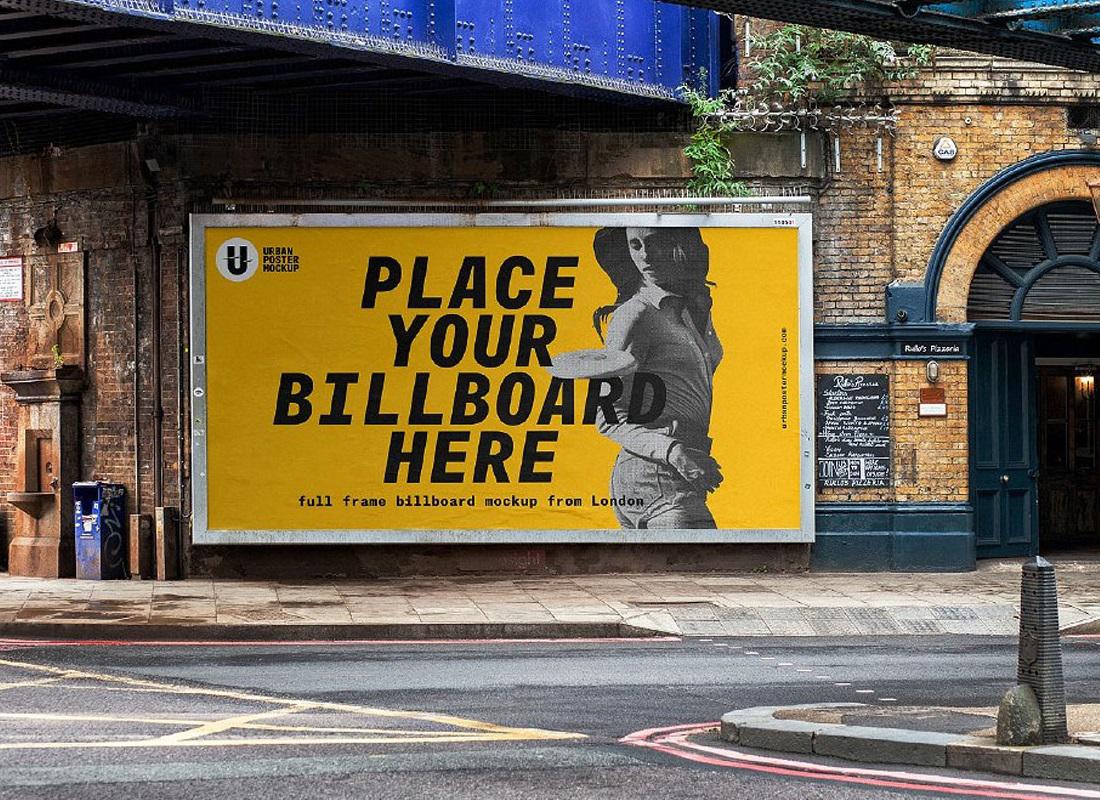}
  \end{subfigure}%
  \begin{subfigure}{.5\textwidth}
  \centering
    \includegraphics[width=1\linewidth]{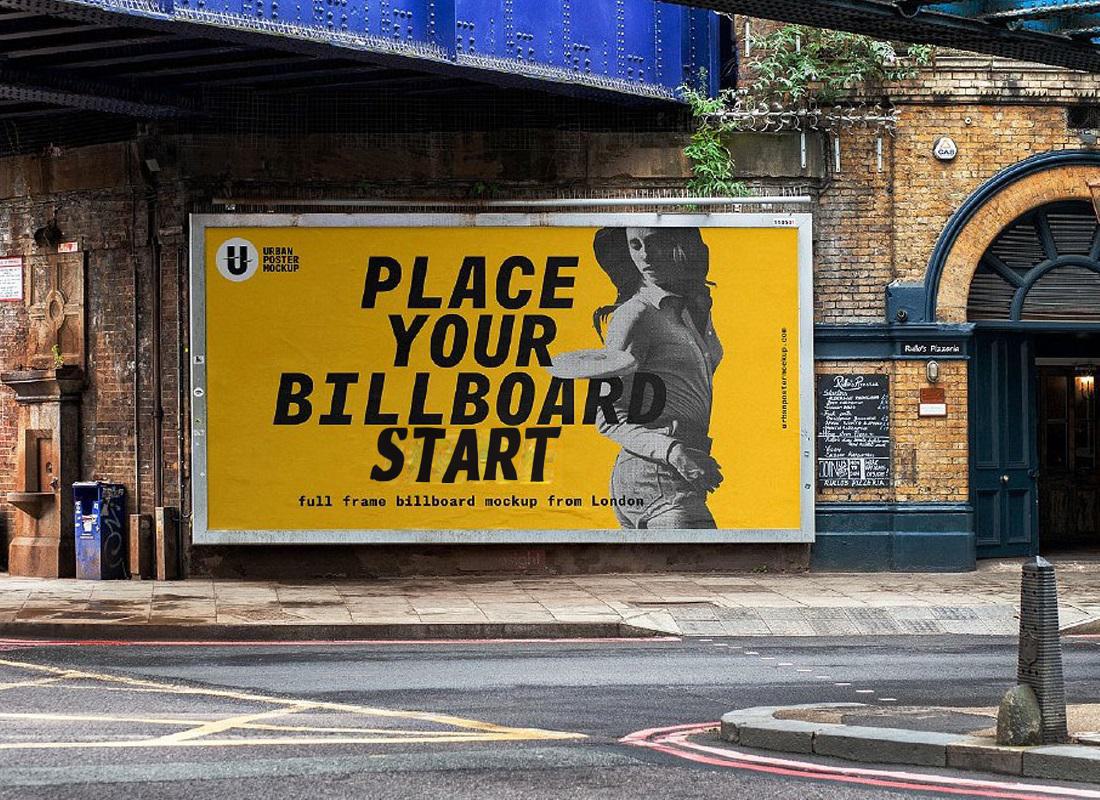}
  \end{subfigure}
  \caption{More Results from Internet Results}
  \label{fig: second_case}
\end{figure*}

Retailers can update prices, product descriptions, and availability information on shelf labels and in online product images, ensuring accuracy and compliance with marketing strategies. One of the best examples as shown in Figure~\ref{fig: second_case} is the "DOUBLEMINT" to "SINGLEMINT" label change with the product brand name. This shows how FASTER can easily impact marketing e-commerce brand visuals. 

\paragraph{\textbf{Artistic and Creative Expressions:}}

As shown in first example in Figure~\ref{fig: first_case}, FASTER can mimick and adapt specific font colors seamlessly with content editing. "HAPPY BIRTHDAY" could integrate "MERRY BIRTHDAY" preserving all kinds of typographic attributes related to font color, style, letter spacing, alignment and font size. This could eventually help artists and designers use FASTER to incorporate text into their visual creations, with unique and expressive design properties in the user-interface. 

\paragraph{\textbf{Augmented Reality (AR) and Mixed Reality (MR):}}

\begin{figure*}[ht]
  \begin{subfigure}{.5\textwidth}
  \centering
    \includegraphics[width=1\linewidth, height = 5.5cm]{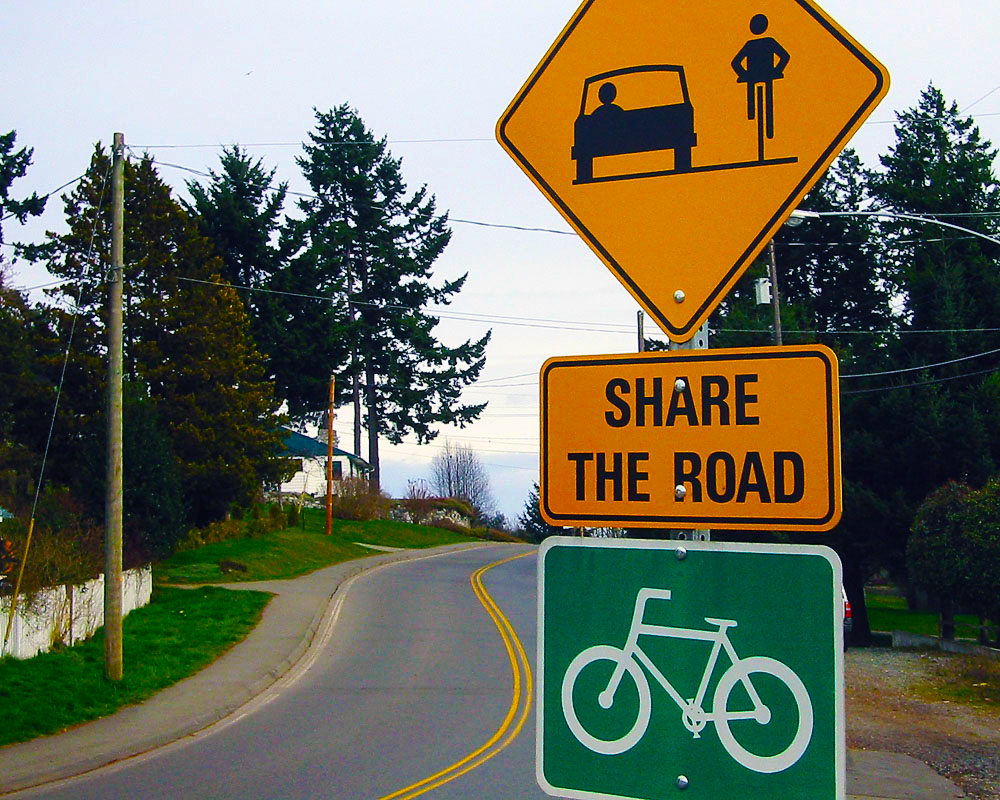}
  \end{subfigure}%
  \begin{subfigure}{.5\textwidth}
  \centering
  \includegraphics[width=1\linewidth, height = 5.5cm]{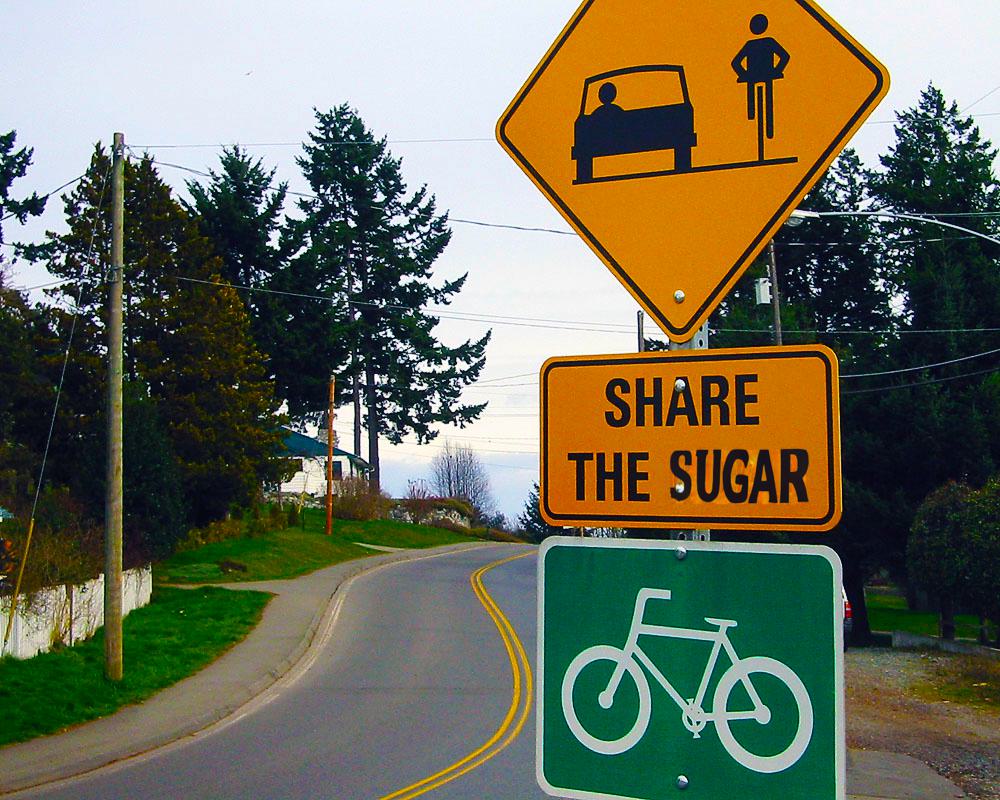}
  \end{subfigure}
  \begin{subfigure}{.5\textwidth}
  \centering
    \includegraphics[width=1\linewidth, height = 5.5cm]{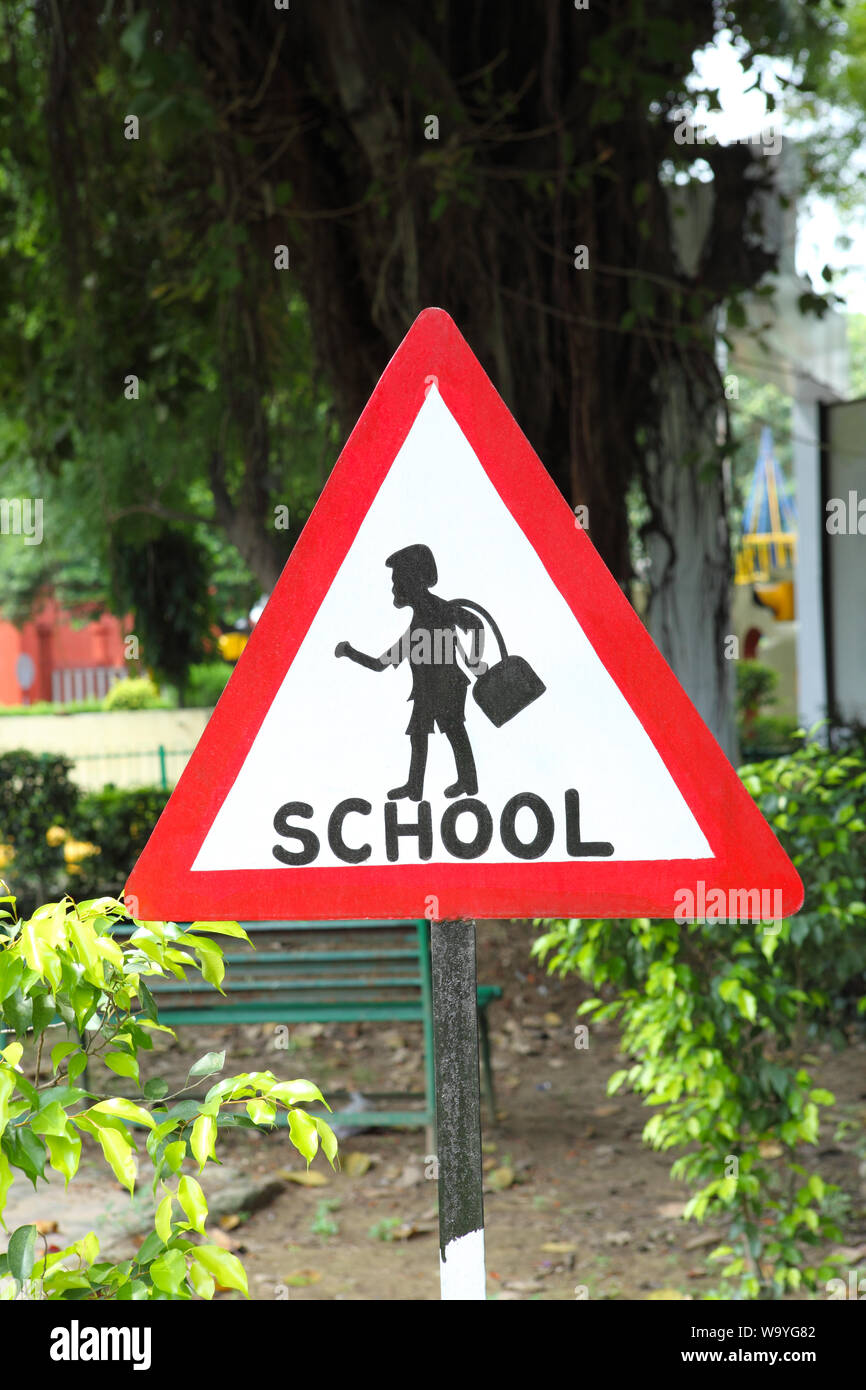}
  \end{subfigure}%
  \begin{subfigure}{.5\textwidth}
  \centering
    \includegraphics[width=1\linewidth, height = 5.5cm]{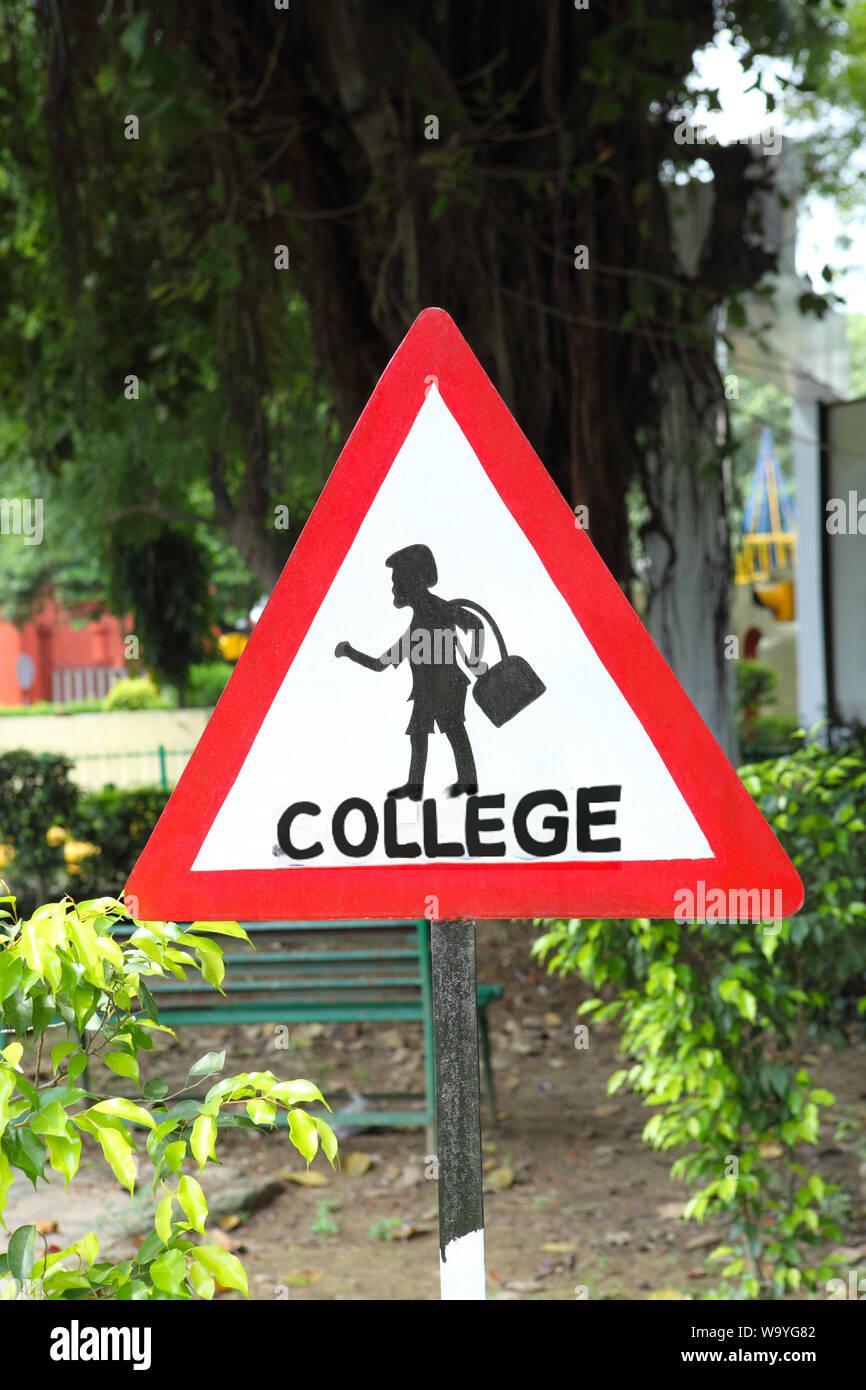}
  \end{subfigure}
  \begin{subfigure}{.5\textwidth}
  \centering
    \includegraphics[width=1\linewidth, height = 7.5cm]{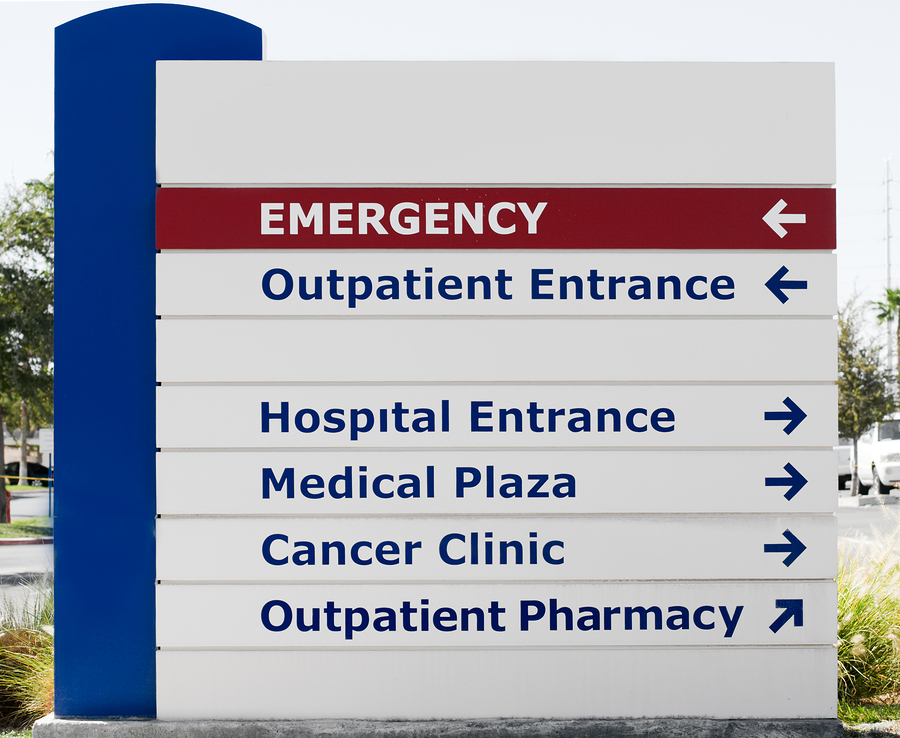}
  \end{subfigure}%
  \begin{subfigure}{.5\textwidth}
  \centering
    \includegraphics[width=1\linewidth, height = 7.5cm]{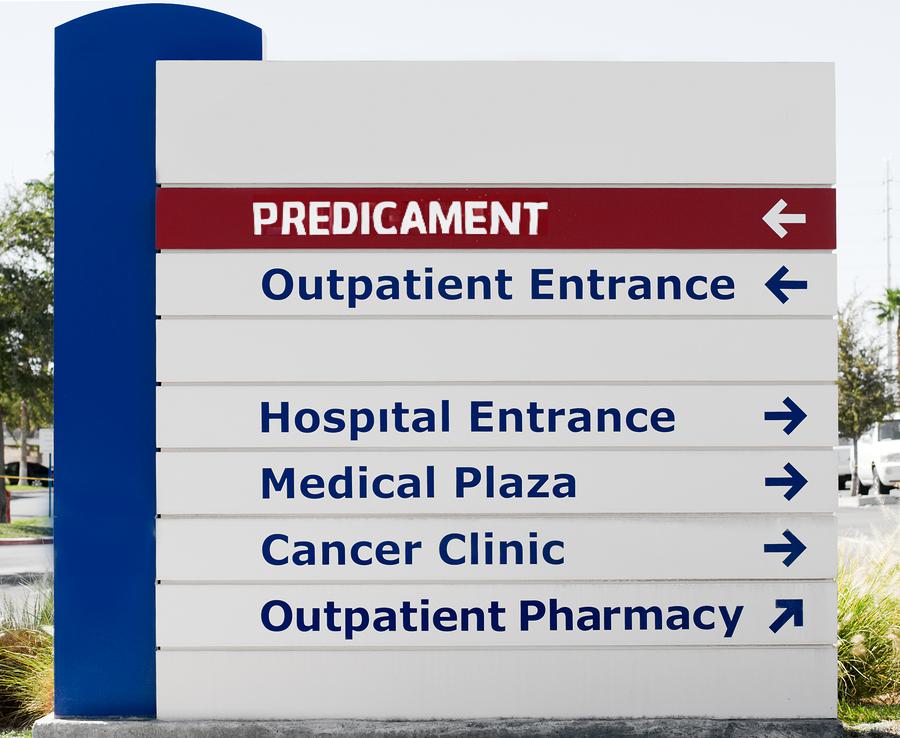}
  \end{subfigure}
  \caption{How FASTER behaves in image editing with stop signals and bulletins in roads}
  \label{fig:fourth_case}
\end{figure*}

FASTER enhances the user experience by overlaying informative text, such as navigation instructions or facts, onto real-world scenes in AR and MR applications. As shown in Figure~\ref{fig:fourth_case}, we observe in all the examples how the content could have tampered with the original navigation instructions in response to street signboards or location milestone blocks. It also promotes visual consistency which ensures that text edits align with the overall context and design, promoting an extremely cohesive user experience.

\paragraph{\textbf{Visually-Rich Document Editing:}} 

\begin{figure*}[ht]
  \begin{subfigure}{.5\textwidth}
  \centering
    \includegraphics[width=1\linewidth, height = 6cm]{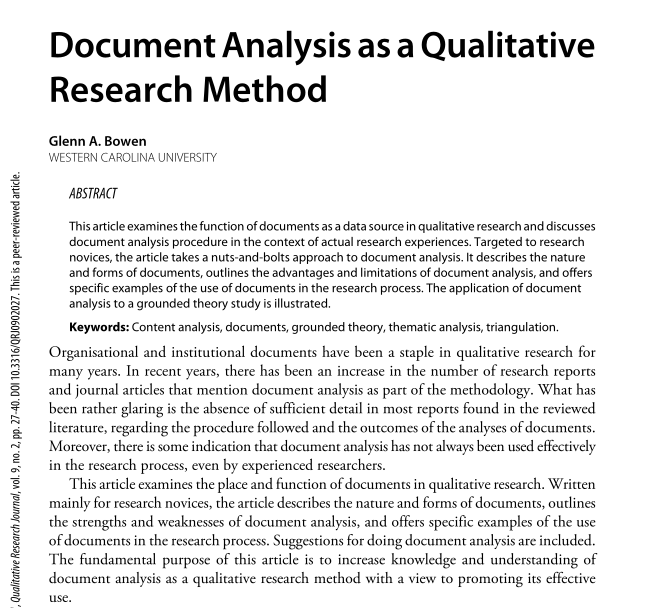}
  \end{subfigure}%
  \begin{subfigure}{.5\textwidth}
  \centering
    \includegraphics[width=1\linewidth, height = 6cm]{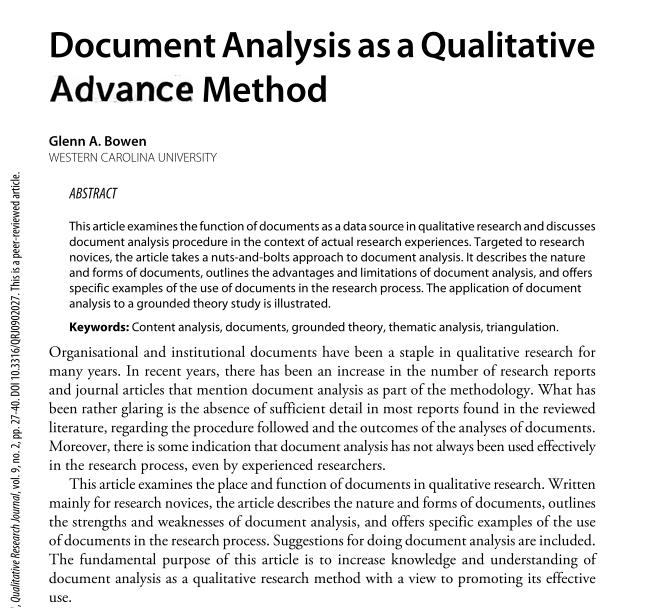}
  \end{subfigure}
  \begin{subfigure}{.5\textwidth}
  \centering
    \includegraphics[width=1\linewidth]{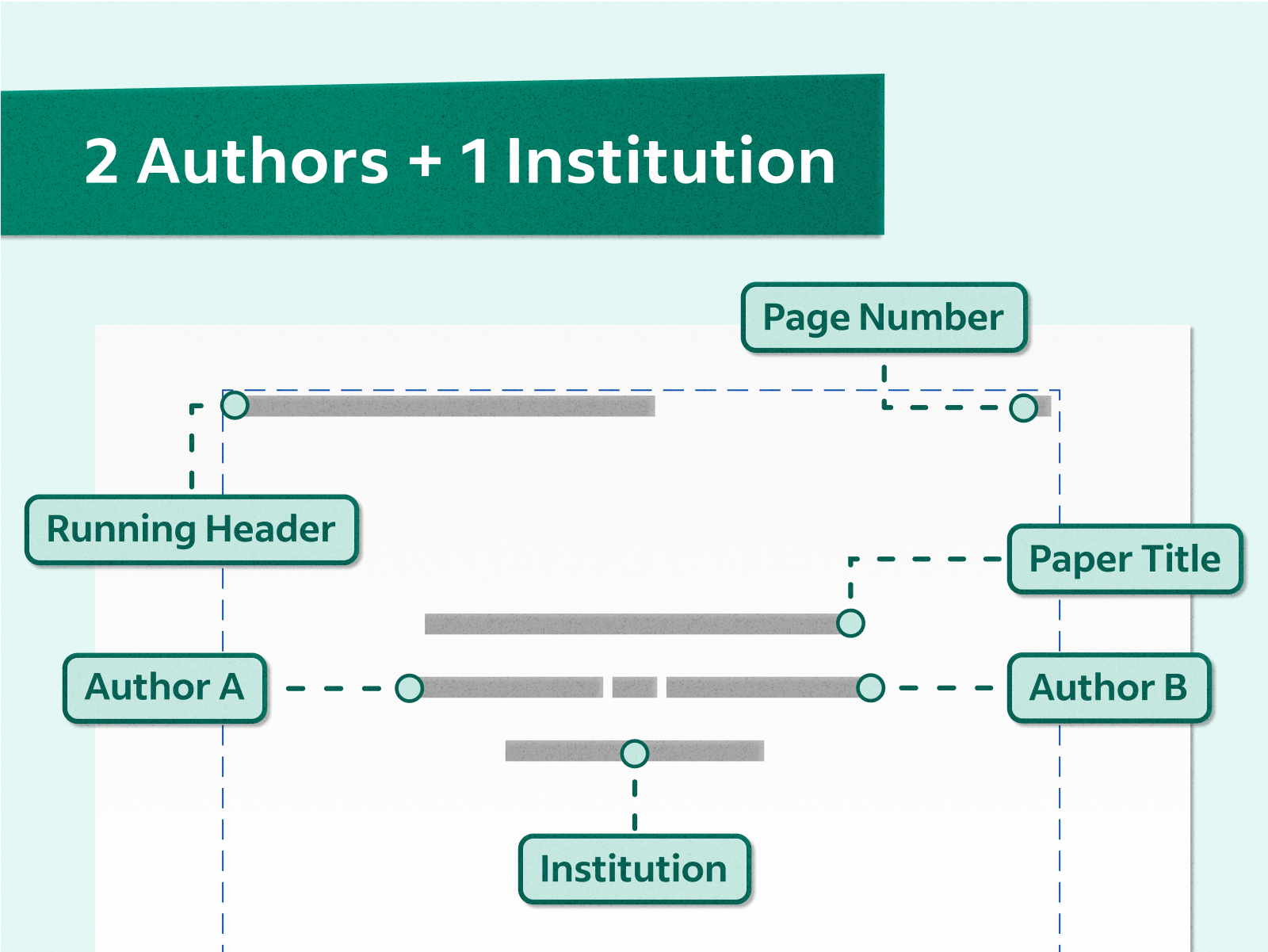}
  \end{subfigure}%
  \begin{subfigure}{.5\textwidth}
  \centering
    \includegraphics[width=1\linewidth]{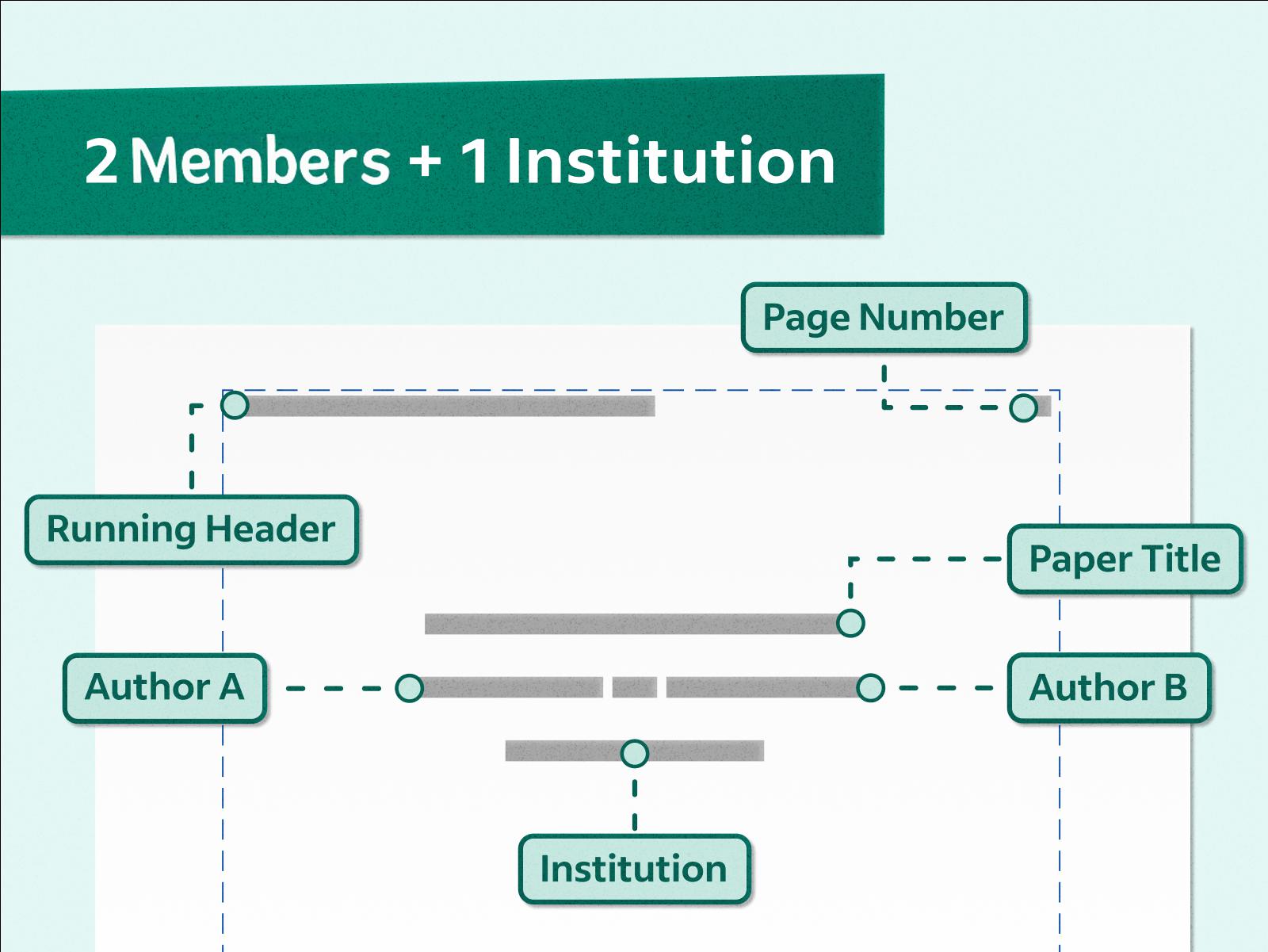}
  \end{subfigure}
  \begin{subfigure}{.5\textwidth}
  \centering
    \includegraphics[width=1\linewidth, height = 8cm]{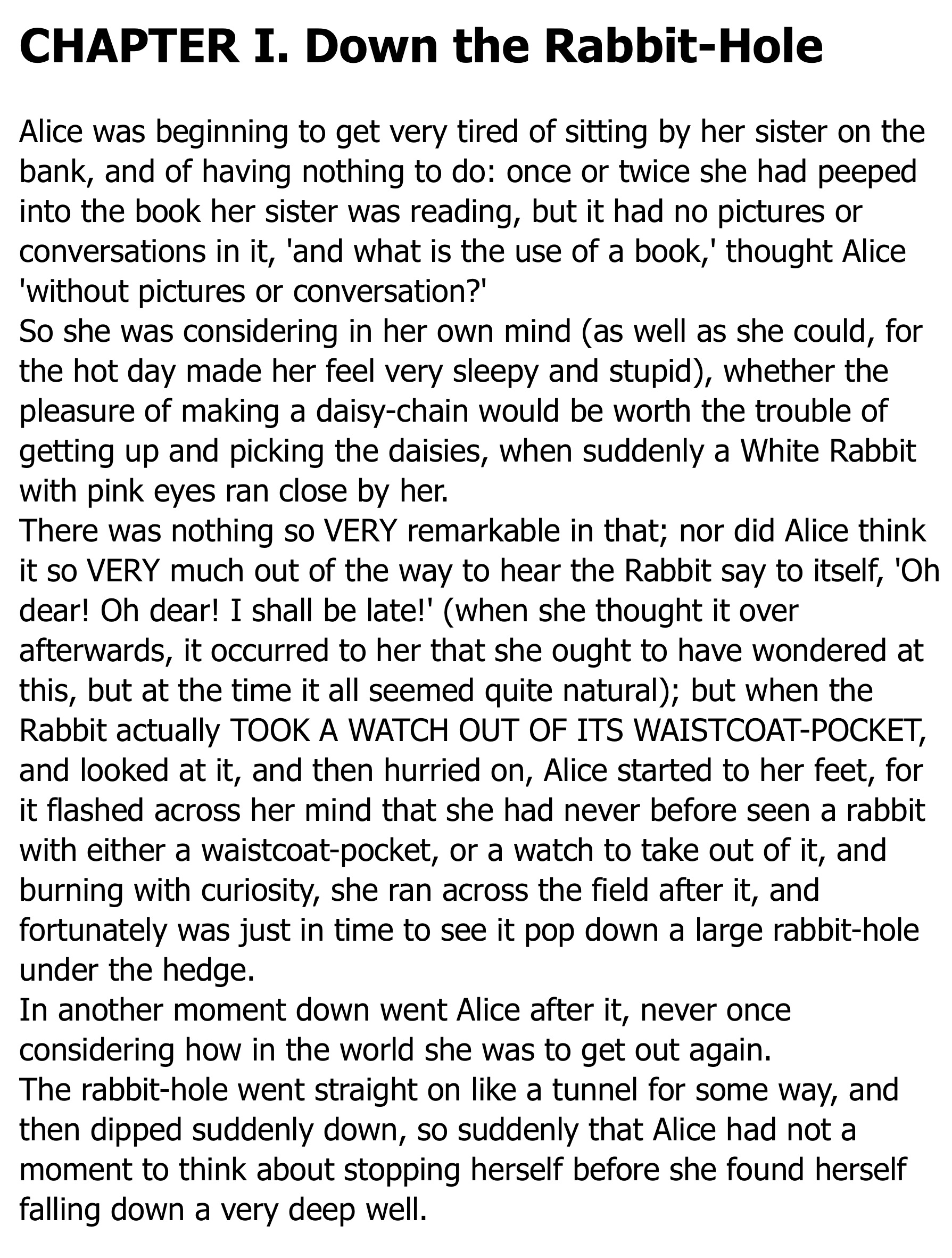}
  \end{subfigure}%
  \begin{subfigure}{.5\textwidth}
  \centering
    \includegraphics[width=1\linewidth, height = 8cm]{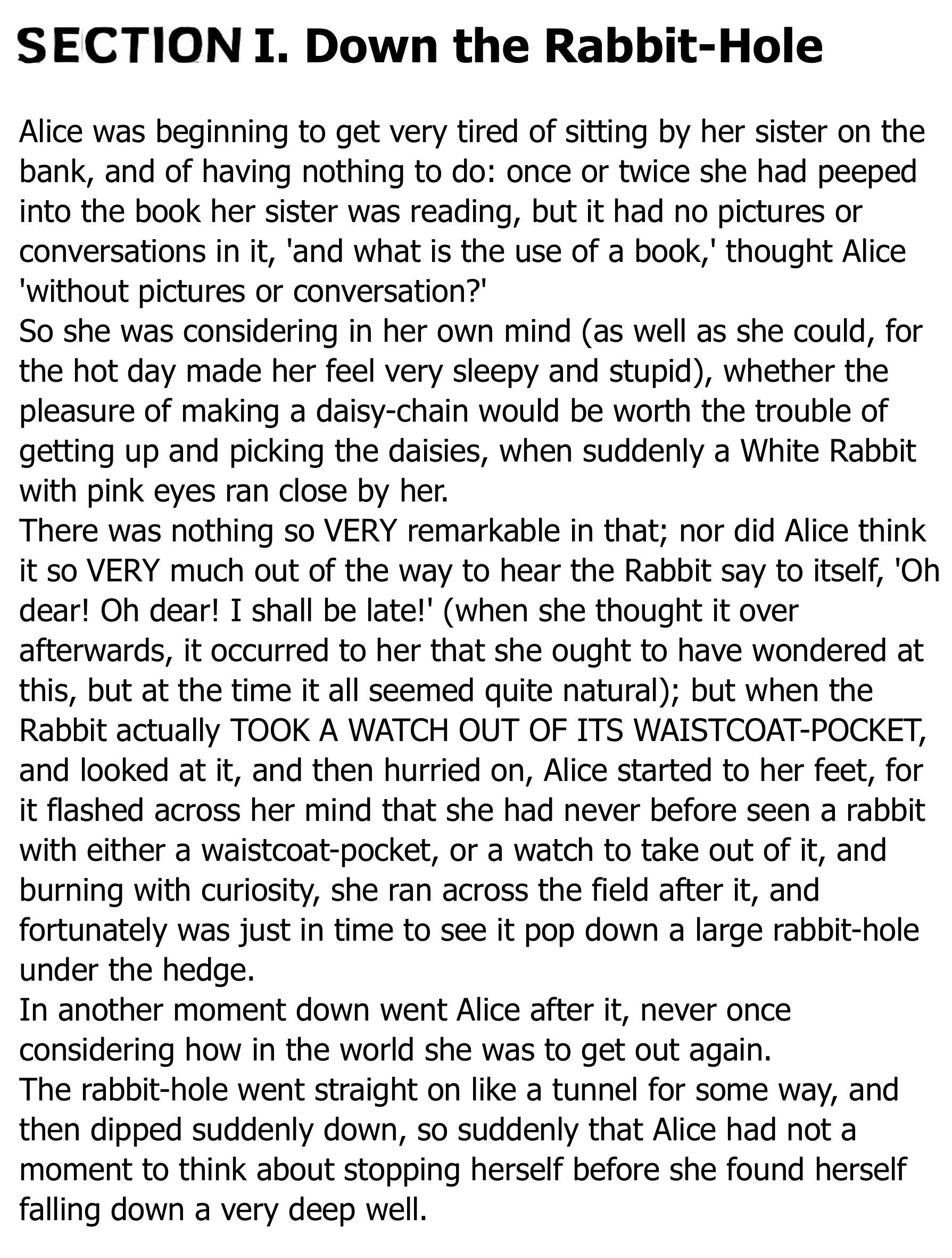}
  \end{subfigure}
  \caption{Some results on image editing with FASTER on Document Images}
  \label{fig:fifth_case}
\end{figure*}

\begin{figure*}[ht]
  \begin{subfigure}{.5\textwidth}
  \centering
    \includegraphics[width=1\linewidth, height = 6cm]{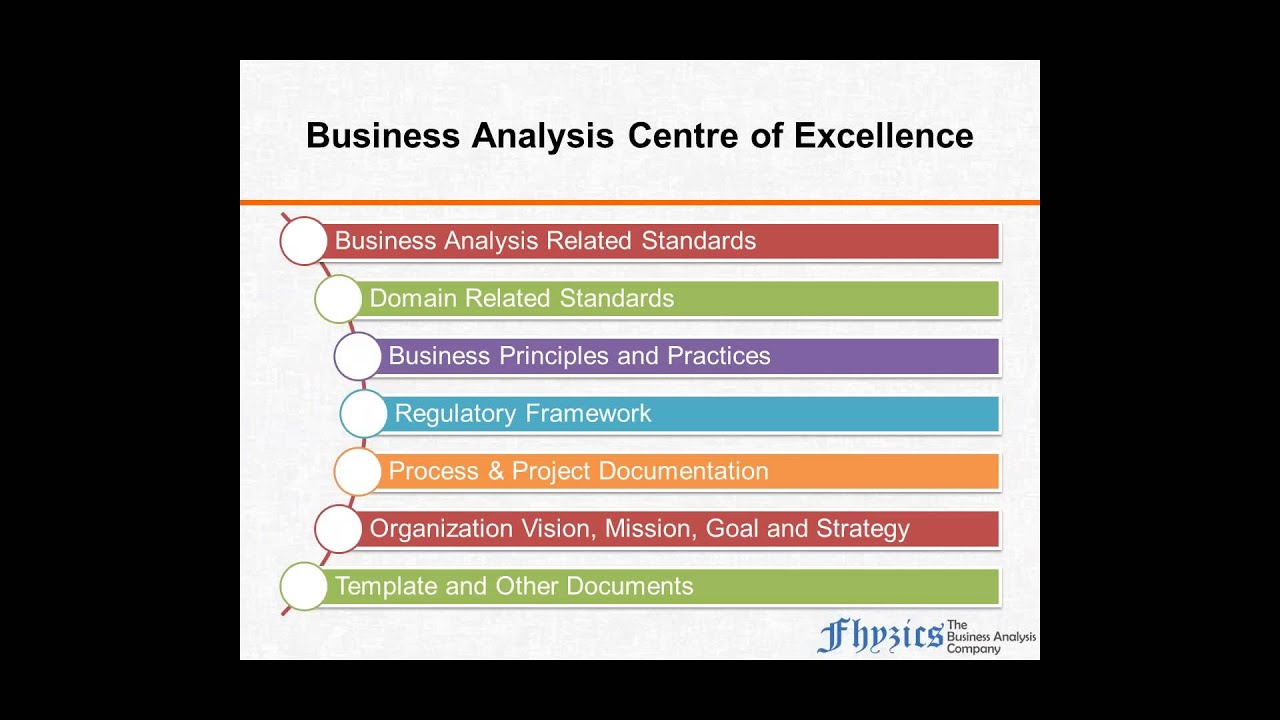}
  \end{subfigure}%
  \begin{subfigure}{.5\textwidth}
  \centering
    \includegraphics[width=1\linewidth, height = 6cm]{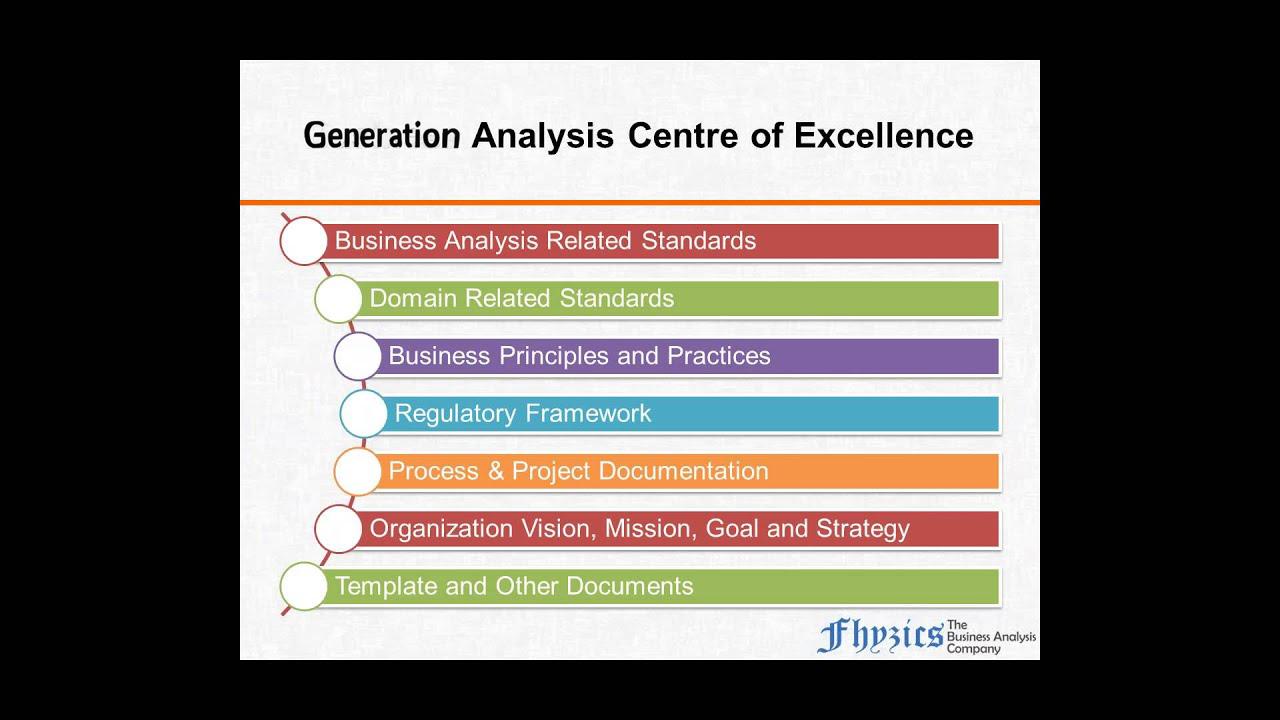}
  \end{subfigure}
  \begin{subfigure}{.5\textwidth}
  \centering
    \includegraphics[width=1\linewidth, height = 6cm]{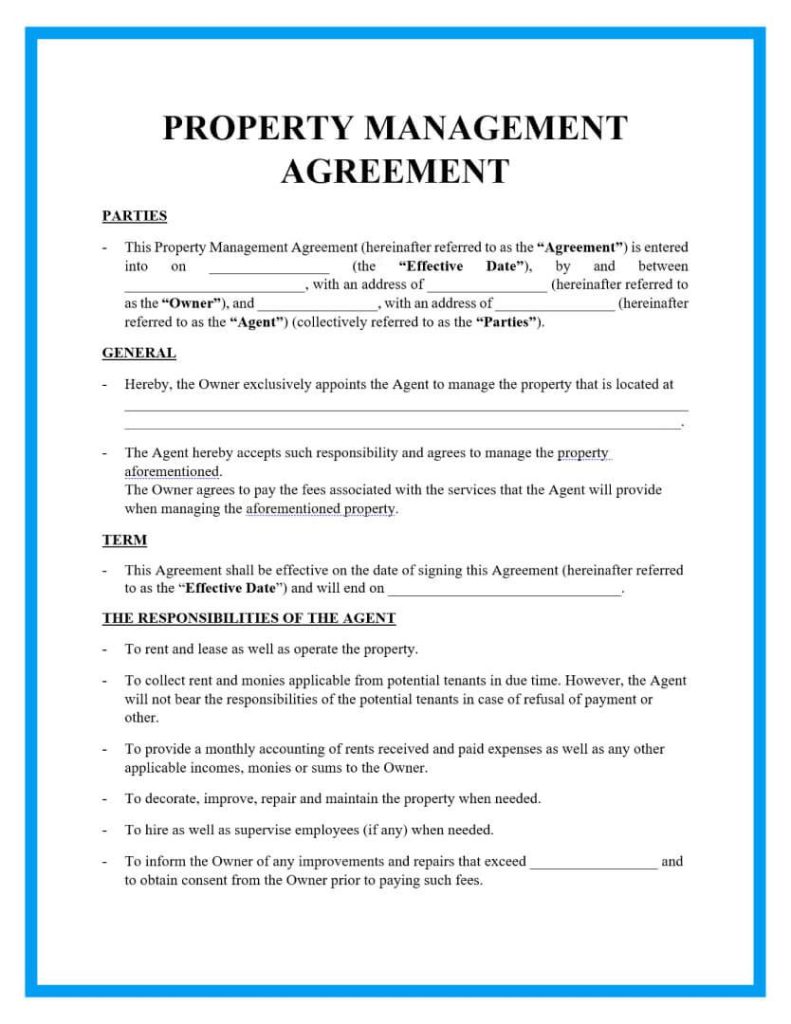}
  \end{subfigure}%
  \begin{subfigure}{.5\textwidth}
  \centering
    \includegraphics[width=1\linewidth, height = 6cm]{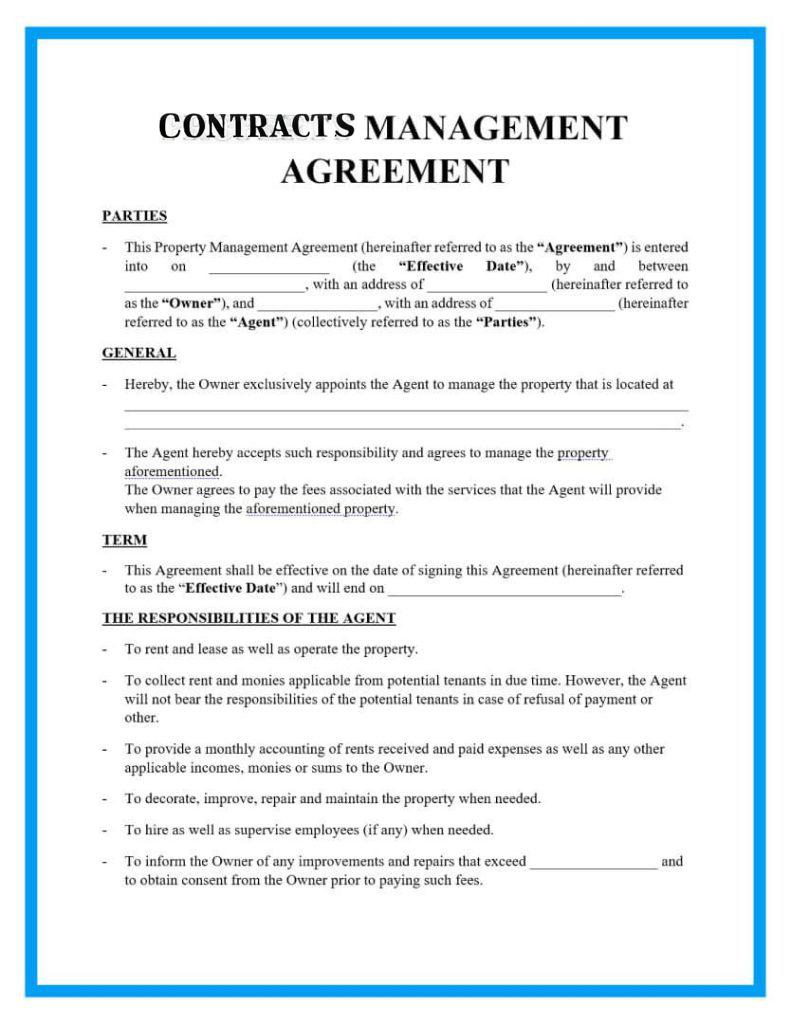}
  \end{subfigure}
  \begin{subfigure}{.5\textwidth}
  \centering
    \includegraphics[width=1\linewidth, height = 7cm]{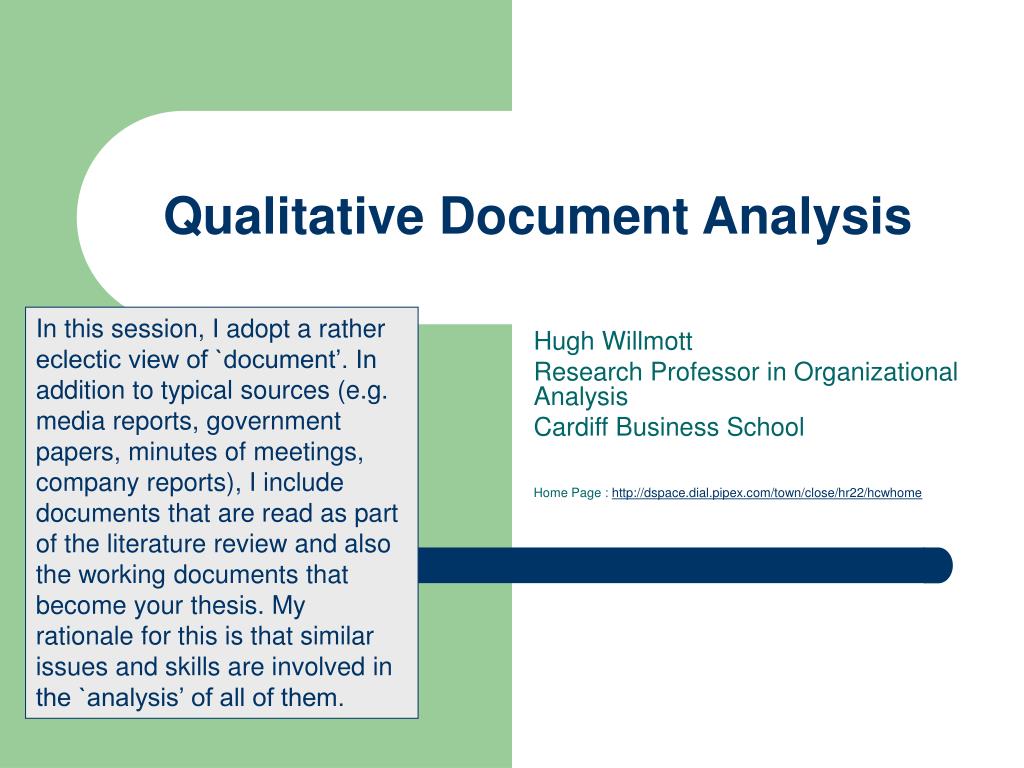}
  \end{subfigure}%
  \begin{subfigure}{.5\textwidth}
  \centering
    \includegraphics[width=1\linewidth, height = 7cm]{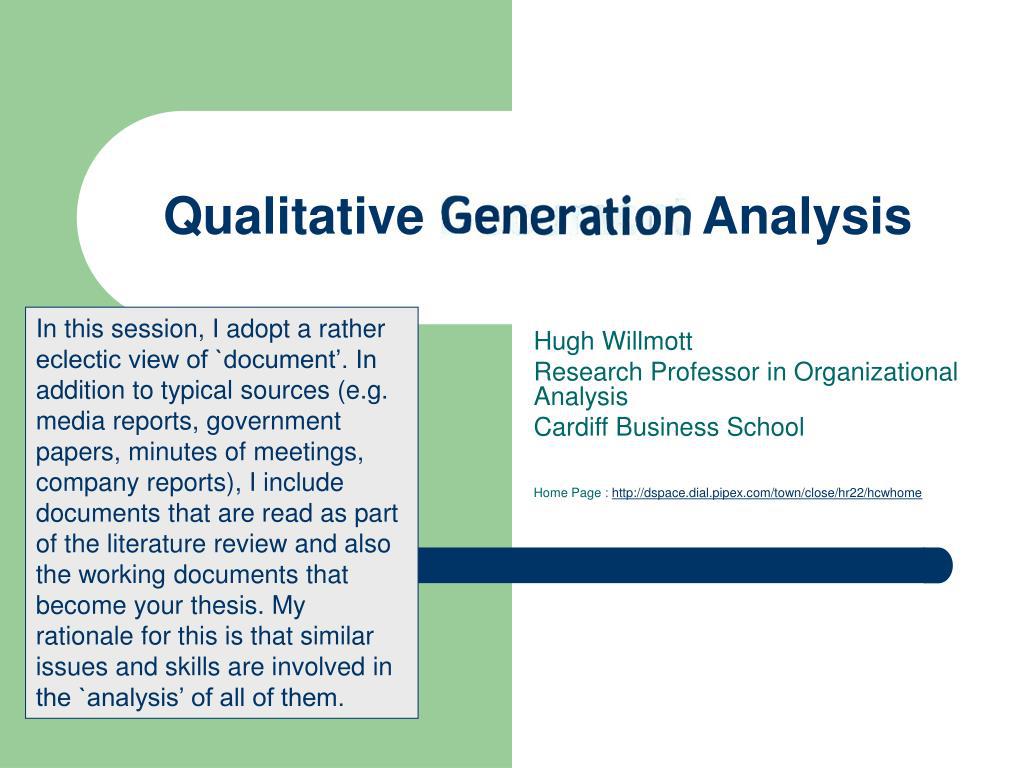}
  \end{subfigure}
  \caption{Some more results on image editing with FASTER on Document Images}
  \label{fig:sixth_case}
\end{figure*}

An interesting case study was performed for some visually-rich document samples as illustrated in Figure~\ref{fig:fifth_case} and Figure~\ref{fig:sixth_case}. The results illustrate that indeed FASTER has some real potential into content editing for scientific papers as shown in Figure~\ref{fig:fifth_case}. This gives it a really powerful application for editing and generative tasks for document analysis. "CHAPTER 1" to "SECTION 1" edit or changing "2 AUTHORS" to "2 Members" in the given document image makes it hard for humans to distinguish between real and synthetically edited (generated) documents. Also, Figure~\ref{fig:sixth_case} shows more infographic-like documents where different header style elements have been preserved with edited content which blends seamlessly with the overall document structure.

\begin{figure*}[ht]
\begin{subfigure}{.5\textwidth}
  \centering
    \includegraphics[width=1\linewidth]{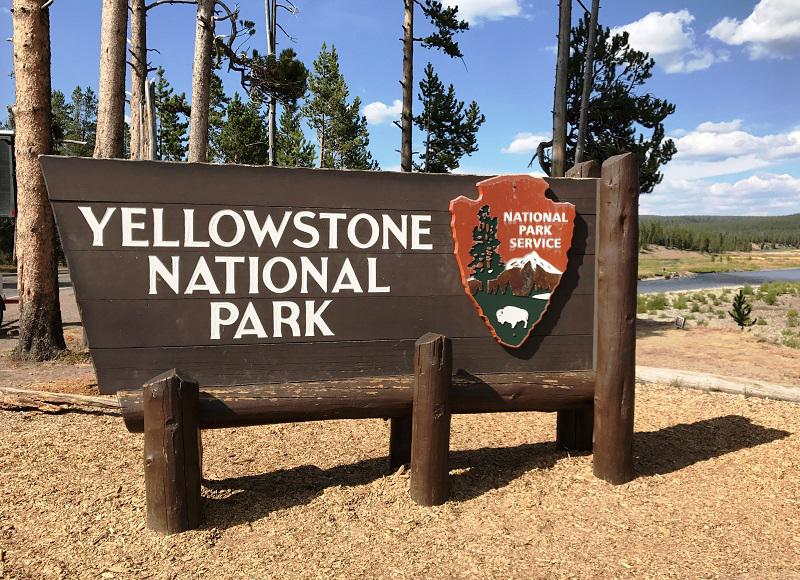}
  \end{subfigure}%
  \begin{subfigure}{.5\textwidth}
  \centering
    \includegraphics[width=1\linewidth]{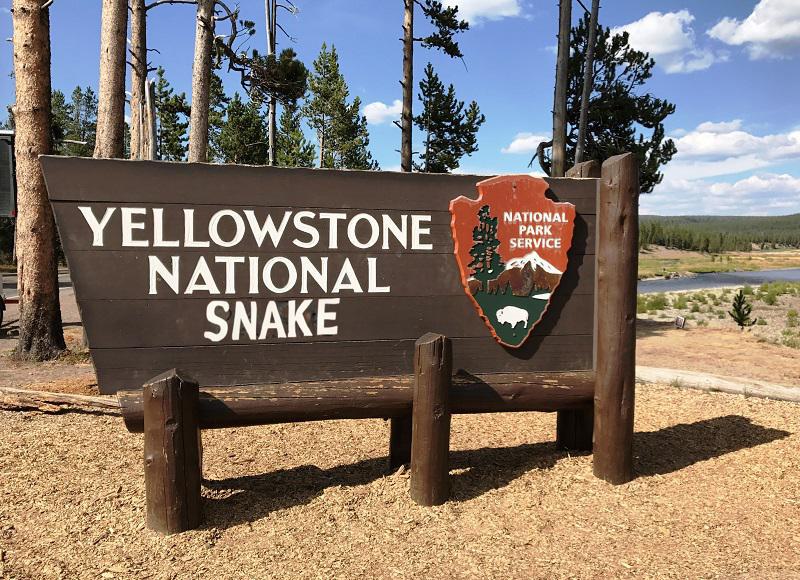}
  \end{subfigure}
  \begin{subfigure}{.5\textwidth}
  \centering
    \includegraphics[width=1\linewidth]{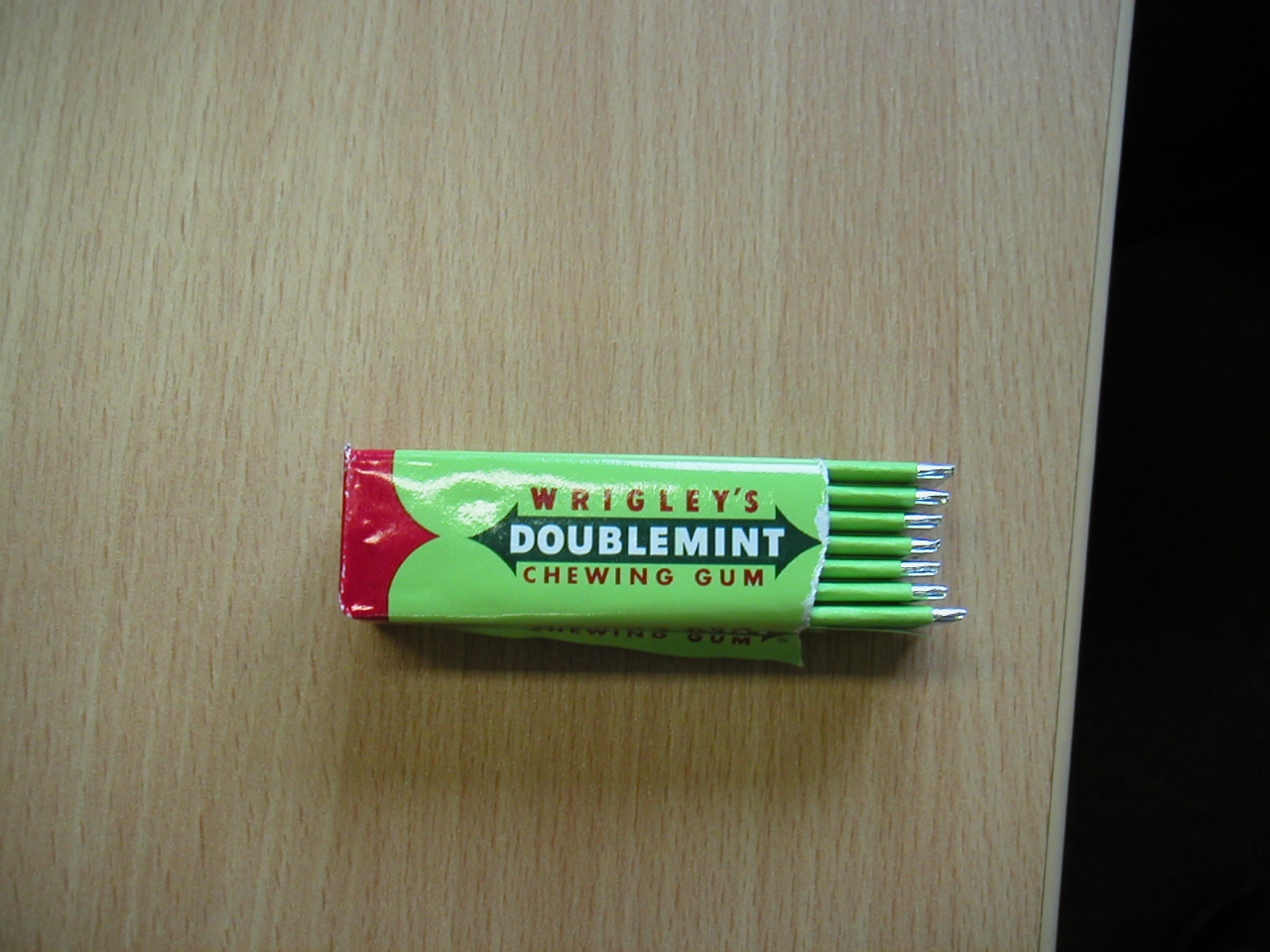}
  \end{subfigure}%
  \begin{subfigure}{.5\textwidth}
  \centering
    \includegraphics[width=1\linewidth]{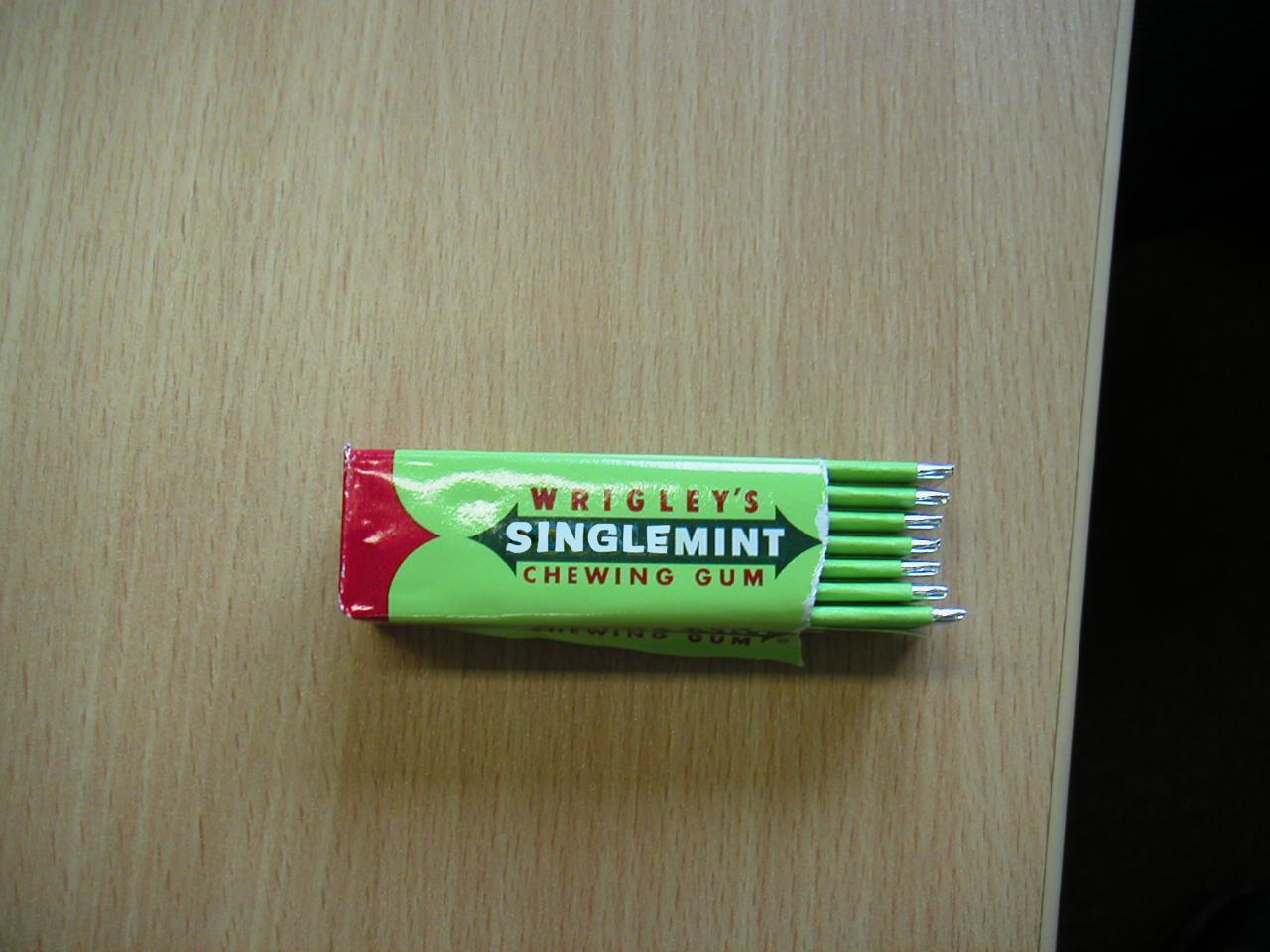}
  \end{subfigure}
  \begin{subfigure}{.5\textwidth}
  \centering
    \includegraphics[width=1\linewidth]{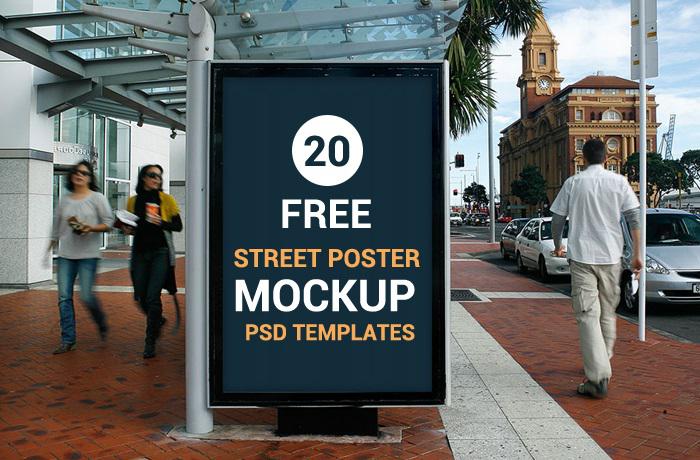}
  \end{subfigure}%
  \begin{subfigure}{.5\textwidth}
  \centering
    \includegraphics[width=1\linewidth]{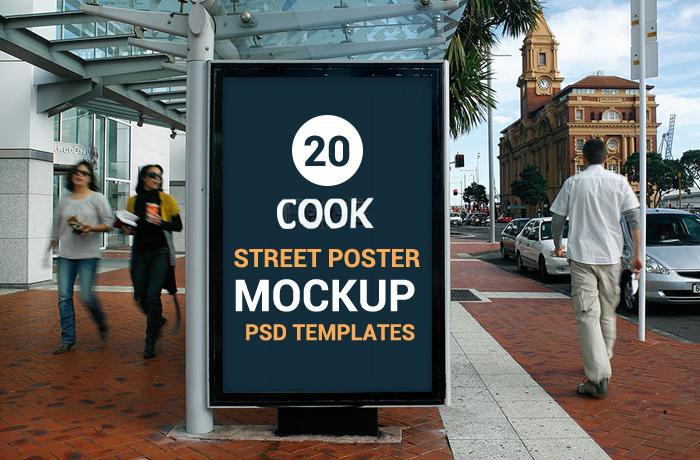}
  \end{subfigure}
  \caption{More Qualitative Results from  real-world Internet images}
\end{figure*}

\begin{figure*}[ht]
  \begin{subfigure}{.5\textwidth}
  \centering
    \includegraphics[width=1\linewidth]{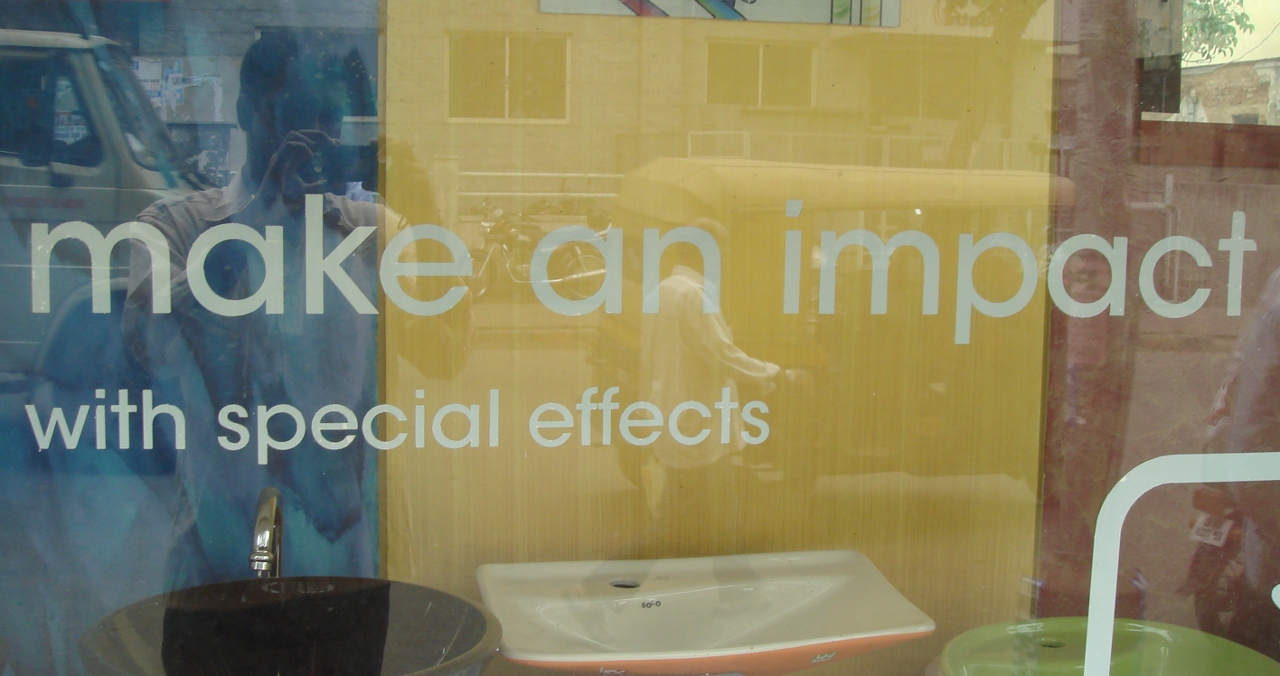}
  \end{subfigure}%
  \begin{subfigure}{.5\textwidth}
  \centering
    \includegraphics[width=1\linewidth]{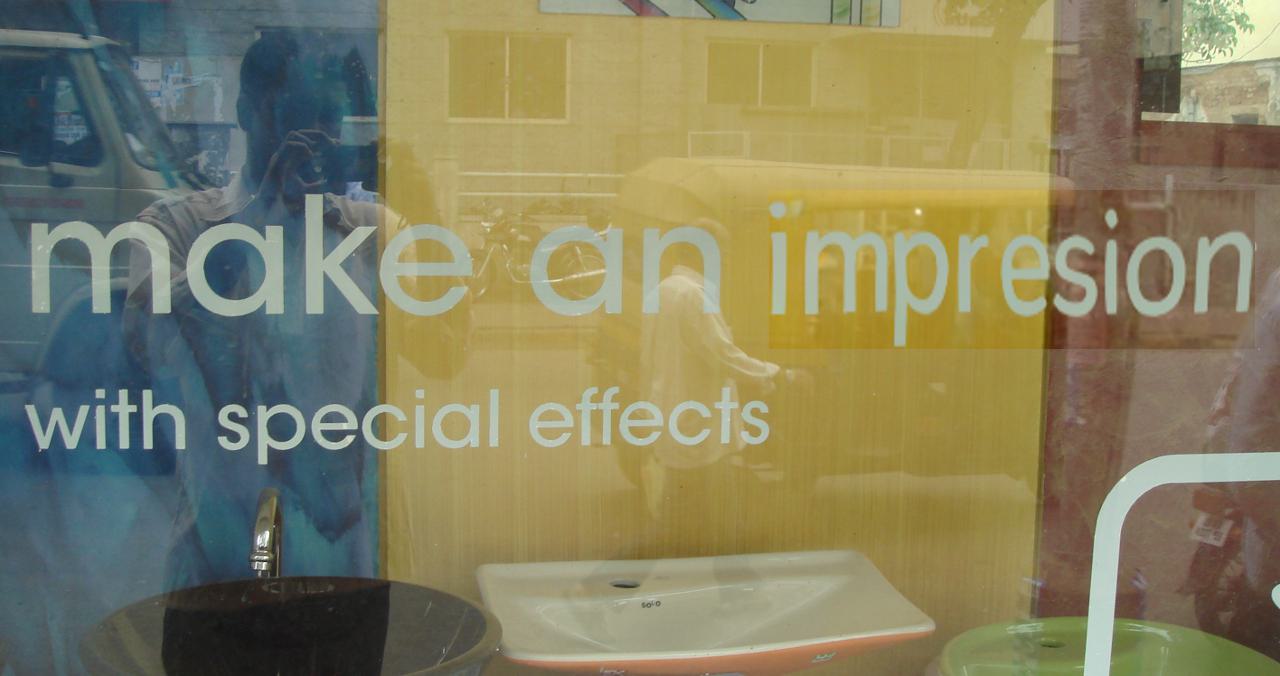}
  \end{subfigure}
  \begin{subfigure}{.5\textwidth}
  \centering
    \includegraphics[width=1\linewidth]{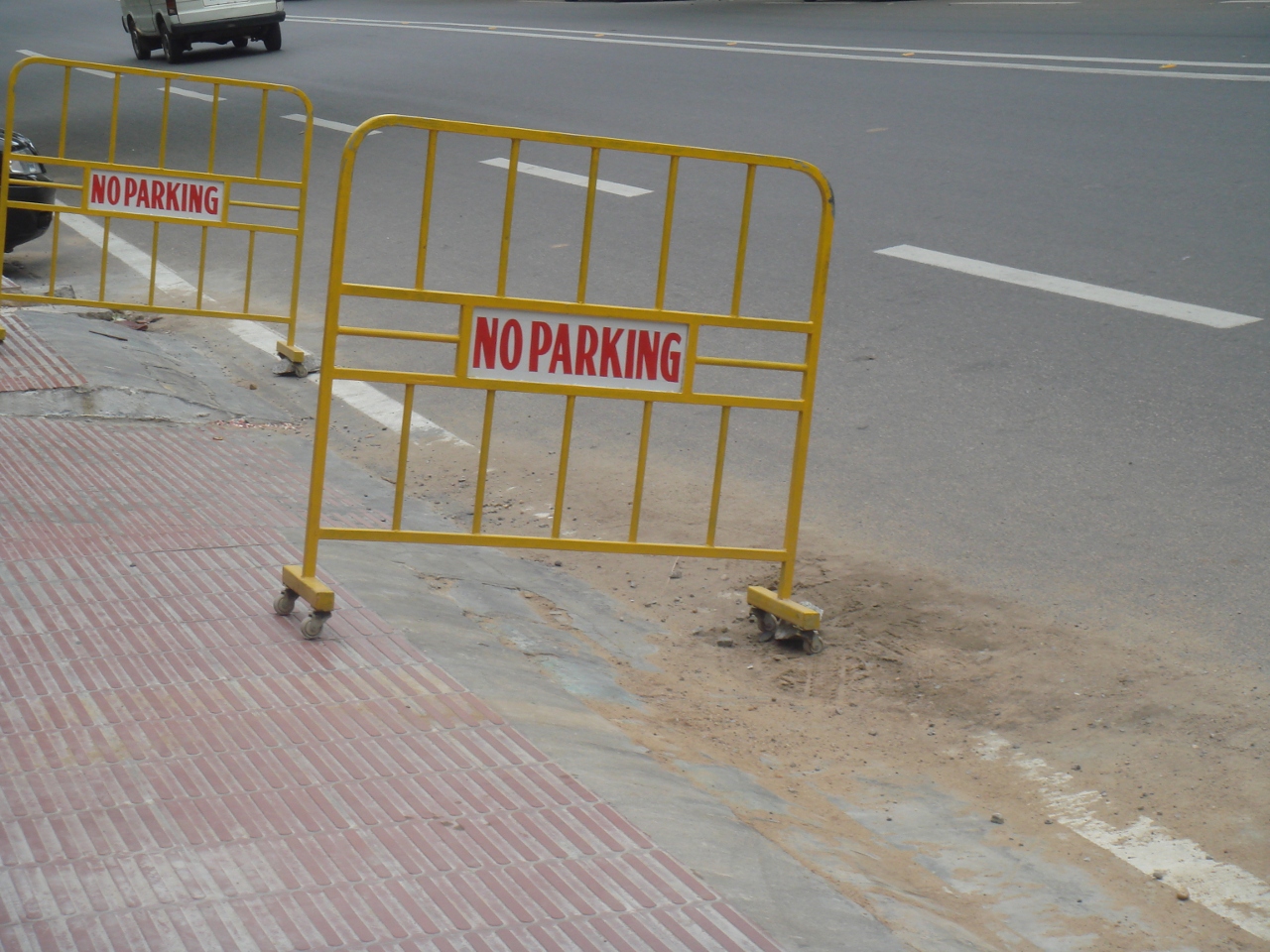}
  \end{subfigure}%
  \begin{subfigure}{.5\textwidth}
  \centering
    \includegraphics[width=1\linewidth]{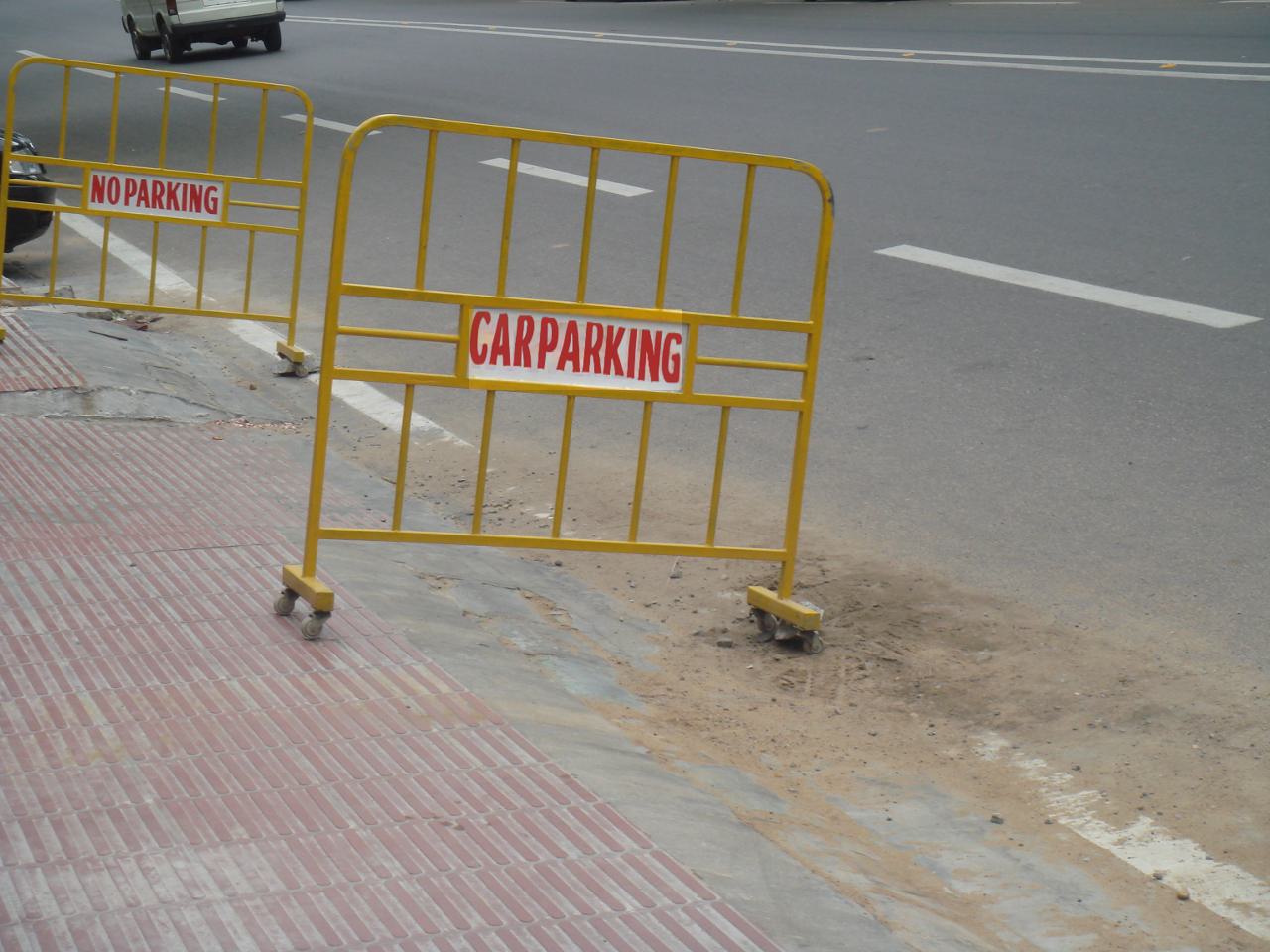}
  \end{subfigure}
  \begin{subfigure}{.5\textwidth}
  \centering
    \includegraphics[width=1\linewidth]{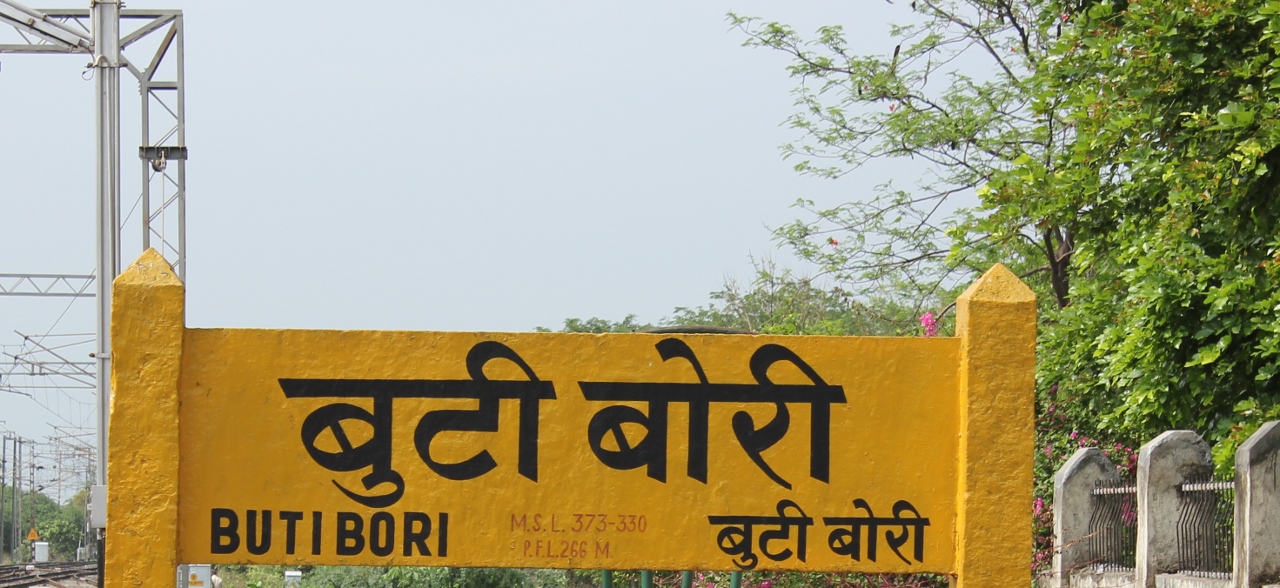}
  \end{subfigure}%
  \begin{subfigure}{.5\textwidth}
  \centering
    \includegraphics[width=1\linewidth]{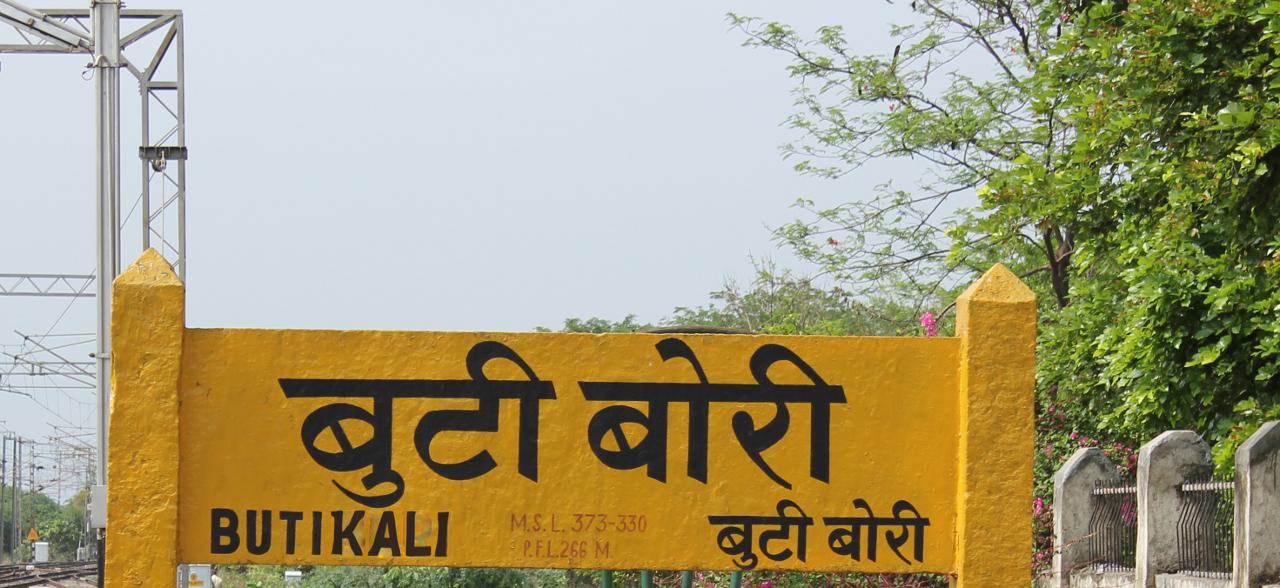}
  \end{subfigure}
  \caption{Some Specific difficult real-world examples}
  \label{fig:third_case}
\end{figure*}

\end{document}